\PassOptionsToPackage{safe}{silence}
\documentclass[10pt,twocolumn,letterpaper]{article}

\usepackage{graphicx}
\usepackage{amsmath}
\usepackage{amssymb}
\usepackage{booktabs}
\usepackage{diagbox}
\usepackage{multirow}
\usepackage{makecell}
\usepackage{tabularx}
\usepackage{marvosym}
\usepackage{verbatim}
\usepackage{colortbl}
\usepackage{xcolor}
\usepackage{svg}
\usepackage{float}
\usepackage{color}
\usepackage[ruled,vlined]{algorithm2e}
\usepackage[pagenumbers]{cvpr} 

\usepackage[ruled,vlined]{algorithm2e}
\usepackage[HTML]{xcolor}
\usepackage{longtable} 
\usepackage{siunitx} 
\usepackage{booktabs} 
\usepackage{xltabular}
\usepackage{listings}
\usepackage[dvipsnames,svgnames]{xcolor}
\usepackage{tikz} 
\usepackage{tkz-euclide}
\usetikzlibrary{shapes.geometric, arrows.meta, positioning, fit, backgrounds, calc, decorations.pathreplacing}

\definecolor{codegreen}{rgb}{0,0.6,0}
\definecolor{codegray}{rgb}{0.5,0.5,0.5}
\definecolor{codepurple}{rgb}{0.58,0,0.82}
\definecolor{backcolour}{rgb}{0.95,0.95,0.92}
\lstdefinestyle{mystyle}{
    backgroundcolor=\color{backcolour},   
    commentstyle=\color{codegreen},
    keywordstyle=\color{magenta},
    numberstyle=\tiny\color{codegray},
    stringstyle=\color{codepurple},
    basicstyle=\ttfamily\footnotesize,
    breakatwhitespace=false,         
    breaklines=true,                 
    captionpos=b,                    
    keepspaces=true,                 
    numbers=left,                    
    numbersep=5pt,                  
    showspaces=false,                
    showstringspaces=false,
    showtabs=false,                  
    tabsize=2
}

\definecolor{cvprblue}{rgb}{0.21,0.49,0.74}
\usepackage[pagebackref,breaklinks,colorlinks,allcolors=cvprblue]{hyperref}

\def\paperID{} 
\def\confName{CVPR}
\def\confYear{2026}

\title{IBISAgent: Reinforcing Pixel-Level Visual Reasoning in MLLMs for Universal Biomedical Object Referring and Segmentation}

\author{Yankai Jiang\textsuperscript{1,2$\bigstar$}, \quad Qiaoru Li\textsuperscript{1*},  \quad Binlu Xu\textsuperscript{1*}, \quad Haoran Sun\textsuperscript{2}, \quad Chao Ding\textsuperscript{2}, \\ 
 \quad Junting Dong\textsuperscript{2}, \quad \textbf{Yuxiang Cai}\textsuperscript{1,3,4\Letter}, \quad Xuhong Zhang\textsuperscript{1,3,4}, \quad Jianwei Yin\textsuperscript{1}\\[6pt]
 \textsuperscript{1}Zhejiang University \quad
 \textsuperscript{2}Shanghai Artificial Intelligence Laboratory \\
 \textsuperscript{3}Ningbo Global Innovation Center, Zhejiang University \\
 \textsuperscript{4}Zhejiang Key Laboratory of Digital-Intelligence Service Technology \\
 \tt\small{jiangyankai@pjlab.org.cn, caiyuxiang@zju.edu.cn}
 }

\begin{document}
\maketitle
\renewcommand{\thefootnote}{\fnsymbol{footnote}}
\footnotetext[1]{Equal contribution. \textsuperscript{$\bigstar$}Project leader. \textsuperscript{\Letter}Corresponding authors.
The codes and trained models are available at \url{https://github.com/Yankai96/IBISAgent}.
}

\begin{abstract}
Recent research on medical MLLMs has shifted its focus from image-level understanding to fine-grained, pixel-level comprehension. Although segmentation serves as the foundation for pixel-level understanding, existing approaches face two major challenges. First, they introduce implicit segmentation tokens and require joint fine-tuning of the MLLM and external pixel decoders, increasing the risk of catastrophic forgetting and limiting out-of-domain generalization. Second, most methods rely on single-pass reasoning and lack the ability to iteratively refine segmentation results. To overcome these limitations, we propose IBISAgent—a novel agentic MLLM that reformulates segmentation as a vision-centric, multi-step decision-making process. IBISAgent enables MLLMs to generate interleaved reasoning and text-based click actions, invoke segmentation tools, and produce high-quality masks without architectural modifications. We also design a two-stage training framework consisting of cold-start SFT and agentic RL with tailored, fine-grained rewards. Through iterative multi-turn visual reasoning, IBISAgent naturally facilitates mask refinement and enhances robustness in complex medical referring and reasoning segmentation tasks.
Extensive experiments demonstrate that IBISAgent consistently outperforms both closed-source and open-source SOTA methods. 
\end{abstract}
    
\section{Introduction}
\label{sec:intro}
\begin{figure}
    \centering
    \includegraphics[width=\linewidth]{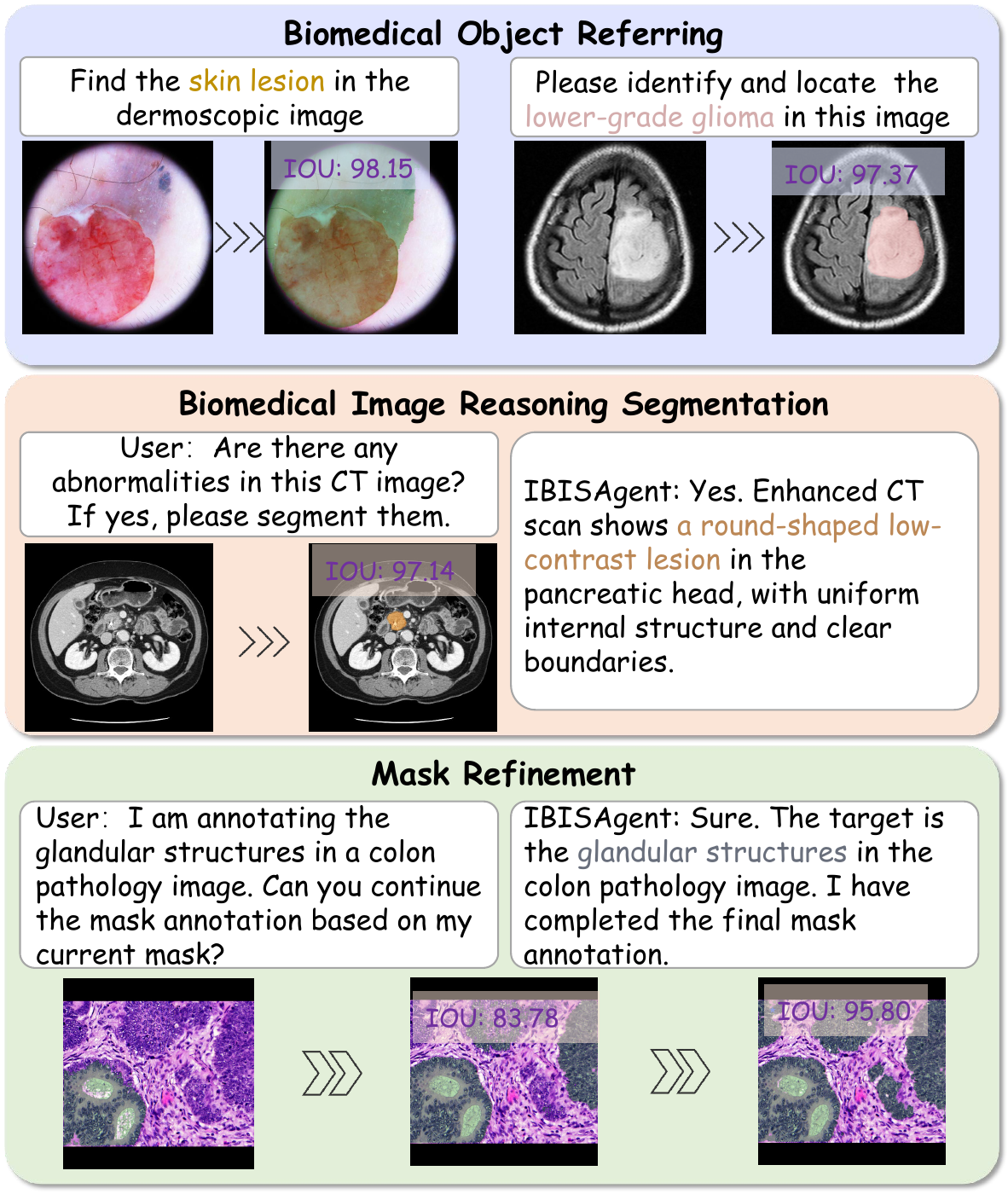}
    
    \caption{
        IBISAgent flexibly supports a wide range of fine-grained biomedical image understanding tasks, including referring and reasoning segmentation. It also handles a novel mask-refinement task that assists annotators in completing partially labeled masks.
    }
    \vspace{-10pt}
    \label{fig:introduction}
\end{figure}
Multimodal large language models (MLLMs) has yielded notable advancements in developing powerful medical AI assistants~\cite{li2023llava,chen2024huatuogpt,lingshu,sellergren2025medgemma}. These models now achieve high scores on exam-style medical question-answering benchmarks
~\cite{lau2018dataset,he2020pathvqa,jin2019pubmedqa,pal2022medmcqa}. However, daily clinical diagnosis is far more complex than a single structured question-and-answer interaction, since holistic medical image analysis comprises multiple subtasks, such as segmentation and detection of biomedical objects~\cite{zhao2024biomedparse}. Prevailing medical MLLMs focus on visual question answering (VQA) and are largely confined to image-level understanding tasks, thus failing to achieve fine-grained pixel-level comprehension.

Recently, increasing efforts have been devoted to equipping MLLMs with pixel-level reasoning segmentation capabilities~\cite{lai2024lisa,yang2023lisa++,bai2024one,ren2024pixellm,liu2025unipixel}. These models typically introduce additional task-specific segmentation tokens (\textit{e}.\textit{g}., \texttt{<SEG>}), which are decoded through external pixel decoders to generate segmentation masks.
Building on this foundation, many studies~\cite{bai2024m3d,wu2025unibiomed,huang2025towards,tong2025medisee} have adopted similar strategies to adapt MLLMs for medical image segmentation. 
Despite their effectiveness, the segmentation capabilities of existing medical MLLMs are still limited. Their dependence on joint fine-tuning of the MLLM and external pixel decoders heightens the risk of \textit{catastrophic forgetting}, resulting in strong in-domain performance but weak cross-domain generalization. Moreover, the introduction of implicit segmentation tokens disrupts the MLLM’s natural text output space, thereby weakening its reasoning ability and failing to reflect the model’s intrinsic pixel-level understanding~\cite{siam2025pixfoundation}.

These limitations motivate a re-examination of how to better elicit pixel-level visual reasoning in MLLMs for biomedical image segmentation (BIS). Unlike natural images, biomedical images often exhibit subtle and complex visual semantics, such as faint lesion cues and nuanced pathological patterns. A single forward pass for segmentation is often inadequate. In contrast, human experts typically perform segmentation in a multi-step, interactive manner. For instance, annotators iteratively refine masks through positive and negative clicks using interactive segmentation tools~\cite{kirillov2023segment,ravi2024sam,Ma_2024}.
It is natural to ask whether current MLLMs can observe an image multiple times, re-evaluate their intermediate decisions, and adapt to feedback to perform self-evolving segmentation—thereby emulating the strategies and interactive behaviors of human annotators through the use of segmentation tools.

Therefore, we propose IBISAgent, a novel agentic MLLM that reformulates segmentation as a multi-step Markov Decision Process. IBISAgent decouples pixel-level visual grounding and mask prediction. It iteratively generates interleaved reasoning and text-based click commands, invokes segmentation tools, and refines the current segmentation based on evolving visual features. 
Compared with previous medical MLLMs that rely on implicit tokens and additional pixel decoders, IBISAgent preserves the LLM’s inherent internal language representations and extends segmentation beyond mere pixel prediction to encompass fine-grained visual reasoning and action planning. This design facilitates vision-centric, multi-step decision-making and enables genuine reasoning to support advanced tasks such as automatic mask refinement (\cref{fig:introduction}). Moreover, by treating segmentation models as plug-and-play tools controllable through language, IBISAgent enhances flexibility and extensibility by eliminating rigidly defined input–output templates (\textit{e}.\textit{g}., ``It’s \texttt{<SEG>}.''), thereby facilitating the implementation of a unified framework across diverse tasks.

To develop IBISAgent, we first employ an automated iterative click-simulation algorithm to transform existing BIS datasets into annotation trajectories. 
Then, we construct a high-quality dataset comprising $400K+$ samples, each annotated with step-wise reasoning traces, click action trajectories, and corresponding masks. We further propose a two-stage training protocol. First, we perform cold-start supervised fine-tuning (SFT) to enable IBISAgent to learn visual reasoning and plan click actions according to user instructions. 
Then we employ reinforcement learning (RL) with novel fine-grained rewards—particularly, a region-based click placement reward and a progressive segmentation improvement reward—to further enhance decision-making, 
allowing the model to autonomously discover efficient and advanced action policies rather than merely imitating the click trajectories learned during the SFT stage.

We conduct a comprehensive evaluation on multiple benchmarks, spanning both in-domain and zero-shot scenarios, to rigorously assess the performance of IBISAgent. Results show that our model exhibits strong pixel-level visual reasoning abilities and significantly exceeds the SOTA MLLMs. Our contributions can be summarized as follows:
\begin{itemize}
    \item We present IBISAgent, a novel agentic framework that equips MLLMs with fine-grained pixel-level visual reasoning, enabling high-quality segmentation without architectural modifications or implicit tokens. 
    \item We construct a comprehensive dataset and introduce effective training strategies—cold-start SFT and RL with tailored fine-grained rewards—to foster strong decision-making and advanced action planning.
    \item We conduct extensive held-in and held-out experiments to evaluate IBISAgent, and the results verify its superiority on biomedical object referring and segmentation tasks.
\end{itemize}

\section{Related Work}
\label{sec:related_work}

\textbf{Pixel-Level Understanding in Medical MLLMs.}
Recent advancements~\cite{wang2025citrusvadvancingmedicalfoundation, wu2025unibiomed, huang2025multimodallargelanguagemodel} in medical MLLMs have increasingly focused on enhancing the models’ fine-grained, pixel-level understanding to enable accurate detection and segmentation of biomedical structures.
Inspired by the pioneering general-domain model LISA~\cite{lai2024lisa}, which introduced a ``reasoning segmentation'' task to enable models to parse complex, implicit text queries and generate corresponding masks, numerous recent studies \cite{wang2025citrusvadvancingmedicalfoundation, huang2025medsegrreasoningsegmentationmedical, trinh2025prsmedpositionreasoningsegmentation, huang2025multimodallargelanguagemodel} have adapted and extended this paradigm to the medical domain to address its unique challenges.
These LISA-style MLLMs follow an innovative ``embedding-as-mask'' paradigm, in which the hidden-state embeddings of a special \texttt{<SEG>} ~token are projected by the LLM and subsequently decoded into segmentation masks via a vision decoder. 
While promising, these methods require MLLMs to learn task-specific implicit tokens and undergo additional joint fine-tuning with a segmentation decoder. This process disrupts the MLLM’s original text output space and increases the risk of catastrophic forgetting, thereby compromising semantic generalization by deviating from language-based outputs. Moreover, existing methods are limited to single-turn reasoning and grounding, lacking an inherent mechanism for autonomous, self-evolving, multi-step refinement of mask predictions. These challenges represent the key issues our work aims to address.

\noindent \textbf{MLLMs with Segmentation Tools.}
Recent studies~\cite{liu2025seg, huang2025sam, li2024mmedagent, liu2025visionreasoner} have explored activating the intrinsic pixel-level understanding capabilities of MLLMs via RL, enabling the models to generate bounding boxes or point prompts that precisely localize target regions. These spatial coordinates are subsequently passed to SAM as prompts to generate the corresponding segmentation masks. However, these methods are constrained to single-turn reasoning and grounding. In complex real-world scenarios, MLLMs often struggle to precisely localize target regions in a single step. In contrast, we reformulate segmentation as a multi-step Markov Decision Process, enabling the MLLM to perform iterative mask refinement and exhibit self-reflective behavior—capabilities absent in prior work. This formulation effectively mitigates error accumulation and substantially improves performance. Furthermore, during the RL process, we design step-wise rewards that provide segmentation-quality–guided feedback at each iteration, further enhancing the agent’s stability and generalization ability.

\section{Method}
IBISAgent is a unified multimodal agent capable of pixel-space reasoning for biomedical image segmentation and mask refinement by adaptively generating spatial prompts and invoking segmentation tools. The ability is inherited from the model’s native capability of visual grounding and action  planning, and further incentivized and enhanced via end-to-end SFT (\cref{sec3.2}) and RL training (\cref{sec3.3}). 


\begin{figure*}
    \centering
    \includegraphics[width=\textwidth]{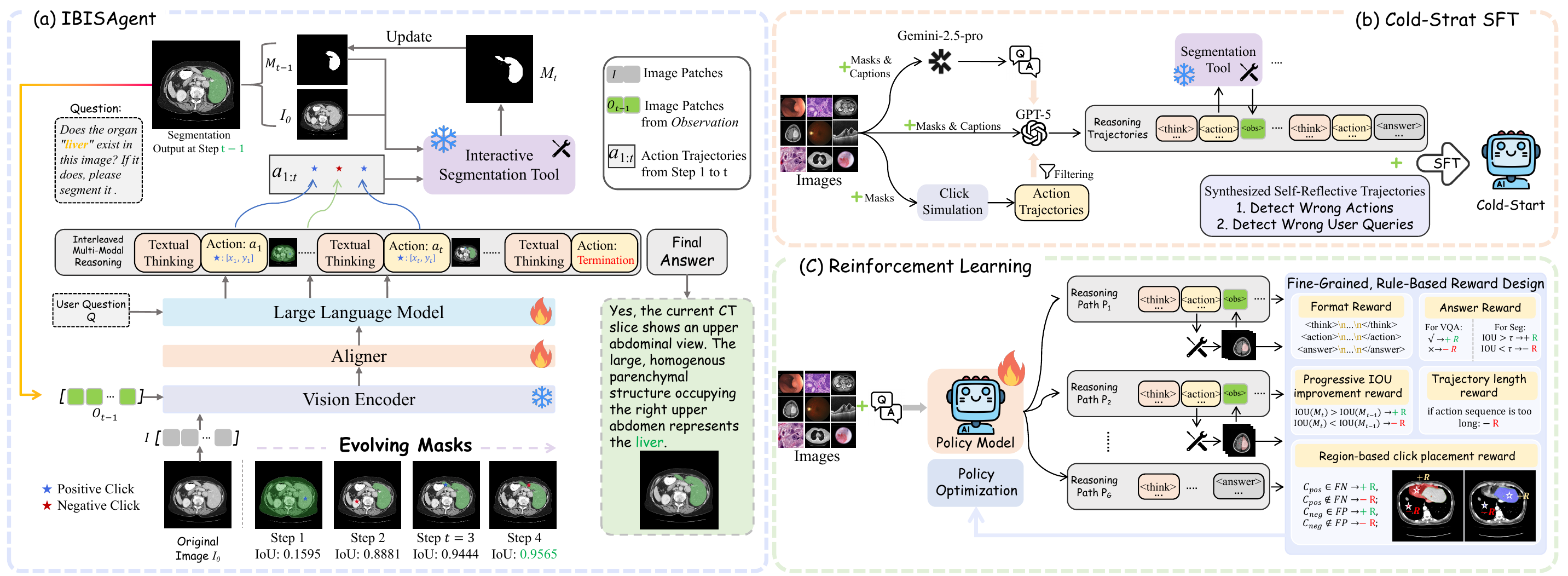}
    
    \caption{
        \textbf{Overview of IBISAgent.} (a) Overall architecture of the agent; (b) illustration of the cold-start SFT training process; and (c) illustration of the RL training process.
    }
    \vspace{-10pt}
    \label{fig:framework}
\end{figure*}

\subsection{Architecture Overview.}
\label{sec3.1}

As illustrated in~\cref{fig:framework}, given a user question $Q$ and an input image $I$, the agent generates
a multi-step, interleaved reasoning path $P$ to derive the final answer. Unlike the predominant textual reasoning paradigm, this pixel-space reasoning path $P$ can be represented as an $T$-step chain: $P = \{(r_t, a_t, o_t)\}^{T}_{t=1}$, where each step comprises textual thinking $r_t$, an action $a_t$ which refers to spatial click operations, and the resulting segmentation observation $o_t$ generated based on $a_t$. This iterative thought-action-observation loop continues until the model reaches a conclusive answer or when predefined limits on context length or interaction turns are reached. The core components are detailed below:
\begin{itemize}
    \item Textual Thinking: The internal reasoning process used by the policy model to select the next action, conditioned on the interaction history and the current observation. 
    \item Action: The action space comprises two options: (1) click operations and (2) emitting a final answer. For click operations, each action is parameterized by three components: \texttt{Target}, which specifies the class name of the current segmentation objective;    
    \texttt{Attribute} $ \in \{+1, -1\}$, which indicates whether the click is positive or negative; and \texttt{Coordinate\_2d} $\in [0, 1]^2$: represents the relative position of the point within the image, scaled to the $[0, 1]$ range for both $x$ and $y$ axes. Note that our method supports performing click operations on multiple targets at once. That is, in the action, it can contain multiple triples consisting of \texttt{Target}, \texttt{Attribute}, and \texttt{Coordinate\_2d}. For clarity, in the following exposition, we focus on the single-target setting as a representative example to describe the method.
    \item Observation: The observation produced by executing $a_t$ in the environment. Concretely, it corresponds to the segmentation mask generated by a segmentation tool (\textit{e}.\textit{g}., MedSAM2~\cite{ma2025medsam2}) given the inputs of the click prompts in $a_t$, the previous step’s output mask $M_{t-1}$ as a spatial prior, and the original image. The initial observation $o_0$ is an empty mask, whereas the final observation represents the optimal target mask. This observation is appended to the interaction history and fed back to the model.
\end{itemize}

\noindent \textbf{Rollout Formulation.} Our objective is to train a policy $\pi_{\theta}$ that emulates the annotation behavior of human experts using an interactive segmentation tool. The policy iteratively generates textual reasoning with a series of positive and negative click points conditioned on the current image $I$ and the evolving mask $M$, progressively refining the segmentation to achieve a high-quality result. 
At each step $t$, the policy is defined as:
\begin{equation}
    r_{t+1}, a_{t+1} \sim \pi_{\theta}\!\left(\cdot \mid I,Q, P_{<t}\right).
\end{equation}

We instruct the model to mark its textual thinking, action, and final answers in the output using the special tokens \texttt{<think>}, \texttt{<action>}, and \texttt{<answer>}. When the model output includes \texttt{<action>}, we automatically parse the action into a format compatible with the interactive segmentation model. 
Subsequently, all previous actions $a_{0:t}$, the current mask $M_{t}$, and the original image $I$ are fed into the interactive segmentation model $F_{seg}$, which generates the updated mask $M_{t+1}$. Furthermore, 
we overlay $M_{t+1}$ as a semi-transparent mask on the original image $I$ to create a new composite image $o_{t+1}$. This image $o_{t+1}$ are then inserted into the \texttt{<obs>} field and
appended to the ongoing trajectory, serving as the input for the MLLM in next step, allowing the model to simultaneously perceive information about $M_{t+1}$ and $I$ in a single frame.
This iterative reasoning process continues until the model determines that the segmentation has reached satisfactory quality. It then outputs a designated end token in \texttt{<action>} and generates the final \texttt{<answer>}.
The system and user prompts are provided in Appendix~\ref{sec:prompts}. Based on this formulation, IBISAgent supports flexible usage scenarios, including both from-scratch segmentation and refinement of pre-existing masks.

\subsection{Cold-Start Supervised Fine-Tuning}
\label{sec3.2}
We aim to cultivate a novel pixel-space reasoning paradigm within IBISAgent. Pure prompting, however, is insufficient to endow multimodal agents with the precision and robustness required to reliably perform iterative visual operations in real-world medical scenarios. To establish a strong foundation for subsequent reinforcement learning, we first initialize IBISAgent’s fine-grained pixel understanding and mask-refinement capability through supervised fine-tuning (SFT) on a cold-start dataset $\mathcal{D}_{\text{cold}}$, which provides ground-truth reasoning and click-annotation trajectories.

\noindent \textbf{Collect Seed Datasets.}
Existing biomedical segmentation datasets contain only final masks without the step-by-step annotation process. Collecting such data by re-hiring human annotators would incur substantial cost. This motivates us to explore whether trajectory supervision can be automatically derived from existing mask annotations using a rule-based procedure. To this end, we utilize the large-scale BiomedParseData~\citep{zhao2024biomedparse}, containing 3.4 million image–mask–label tuples that cover $82$ biomedical object types across $9$ imaging modalities. The dataset provides diverse, high-quality masks that supply rich pixel-level details for fine-grained analysis and serve as reliable supervision for synthesizing click-based annotation trajectories.

\noindent \textbf{Trajectory Generation.}
Inspired by prior research on interactive segmentation, we adopt the click-simulation strategy $F_{\text{cs}}$ proposed in~\cite{xu2016deep}.
Given the current mask $M_t$ and the GT mask $M_{\text{gt}}$, $F_{\text{cs}}$ outputs the next click action $a_{t+1} = F_{\text{cs}}(M_t, M_{\text{gt}})$. Specifically, this function computes the false positive and false negative regions between the current mask $M_t$ and the GT mask $M_{\text{gt}}$, placing the next click action at the center of the error region based on the size and position of these regions. Based on $F_{\text{sim}}$, we can simulate high-quality trajectories $[M_0, a_0, M_1, a_1, \ldots, M_T, a_T]$. 
The full trajectory generation algorithm and an illustrative example are presented in the Appendix~\ref{sec:TrajGen}.

\noindent \textbf{Question, Answer, and Reasoning Generation.}
We first filter the data based on trajectory quality by removing (1) overly long trajectories and (2) those whose final mask Dice score falls below a predefined threshold. For the remaining samples, we prompt Gemini-2.5-Pro~\citep{comanici2025gemini} to generate question–answer pairs conditioned on each image, its GT mask, and the corresponding mask description. This process produces fine-grained queries that explicitly focus on grounding and segmenting the specified mask regions.
The question set $Q$ includes diverse query types, ranging from those that explicitly specify the segmentation target to those that require the model to first reason over the image and adaptively identify the target region on its own. The prompts for QA generation and examples are provided in Appendix~\ref{sec:QAGen}. 
We further use GPT-5~\citep{gpt5chat} to synthesize reasoning for each click action, conditioning on the QA, the correct next action, and pixel-level TP/FP/FN information of the current mask. All generated reasoning traces are post-filtered for format and correctness by human annotations. More details are provided in Appendix~\ref{sec:ReasonGen}. 

\noindent \textbf{Reflective Behavior Synthesis.}
In complex scenarios, if the model cannot backtrack or undo previous actions, an inaccurate decision at any step may propagate and negatively impact subsequent predictions, ultimately degrading the final segmentation quality. To enhance robustness, we synthesize additional samples with self-reflection trajectories, covering two types of error correction:
(1) Self-correction, where the agent detects a wrong action, reverts to the previous state, and re-reasons over the interaction history to produce the correct action; and
(2) User inconsistency correction, where, in mask-refinement scenarios, if the segmentation target described in the instruction does not align with the initial mask, the agent first discards the erroneous mask and re-segments according to the user instruction. 

\noindent \textbf{Cold-Start Training Objective.} Through systematic curation, we obtain a dataset $\mathcal{D}_{\text{cold}}$ containing $456$K samples, including both gold-standard and error-induced self-correction trajectories for cold-start SFT. 
The training objective is to minimize the average negative log-likelihood over all reasoning and action tokens. We employ the standard SFT loss for training. Specifically, we apply loss masks to tokens corresponding to segmentation outputs from executed actions as well as to designated erroneous actions within the self-correction trajectories. Masking the erroneous actions prevents the policy from learning to execute the incorrect actions.
Our SFT strategy endows the model with strong pixel-level reasoning and self-reflection abilities, providing a solid foundation for subsequent RL.

\begin{table*}
\huge
\centering
\renewcommand{\arraystretch}{1.15}
\resizebox{\textwidth}{!}{
\begin{tabular}{l|cccc|cccc|c}
\toprule
\multirow{2}{*}{\diagbox{Benchmarks}{Methods}} &
\multicolumn{4}{c|}{\textcolor{red!48}{\textbf{General-Purpose MLLMs with Segmentation Capability}}} &
\multicolumn{4}{c|}{\textcolor{red!48}{\textbf{Medical MLLMs with Segmentation Capability}}} & \multicolumn{1}{c}{\textbf{Ours}}\\
\cline{2-5}\cline{6-9}\cline{10-10}
& LISA~\cite{lai2024lisa} & LISA++~\cite{yang2023lisa++} & SAM4MLLM~\cite{chen2024sam4mllm} & VisionReasoner~\cite{liu2025visionreasoner} & MedPLIB~\cite{huang2025towards} & Citrus-V~\cite{wang2025citrusvadvancingmedicalfoundation} & UniBiomed~\cite{wu2025unibiomed} & MMedAgent~\cite{li2024mmedagent} & IBISAgent\\
\midrule

\multicolumn{10}{l}{\textbf{In-domain testset $\mathcal{D}_{\text{test}}$}} \\
\midrule
IOU $\uparrow$          & \textcolor{blue}{9.44} (20.46) & \textcolor{blue}{9.49} (20.76) 
                        & \textcolor{blue}{15.85} (27.84) & \textcolor{blue}{16.11} (29.11)
                        
& 22.29 & 30.61 & 50.74 &36.13 &\textbf{85.58}\\
DSC $\uparrow$          & \textcolor{blue}{14.11} (25.73) & \textcolor{blue}{14.30} (25.94) 
                        & \textcolor{blue}{21.16} (33.04)& \textcolor{blue}{22.05} (35.50)

& 27.35 & 37.63 & 58.31 &42.85 &\textbf{92.21}\\
F1-score $\uparrow$     & \textcolor{blue}{20.18} (32.15) & \textcolor{blue}{20.75} (32.34) 
                        & \textcolor{blue}{32.53} (42.75)& \textcolor{blue}{34.78} (46.72)

& 38.94 & 53.75 & 69.22 &56.64 &\textbf{96.39}\\
\midrule

\multicolumn{10}{l}{\textbf{Out-of-domain testset MeCOVQA-G+}} \\
\midrule
IOU $\uparrow$       & \textcolor{blue}{10.07} (15.24) & \textcolor{blue}{9.87} (15.01)
                     & \textcolor{blue}{16.99} (21.19) & \textcolor{blue}{18.27} (24.46)
                     
& 33.36 & 46.54 & 24.88  & 26.54 & \textbf{80.63}\\
DSC $\uparrow$          & \textcolor{blue}{15.44} (21.30) & \textcolor{blue}{14.70} (21.26)
                        & \textcolor{blue}{21.85} (26.35)& \textcolor{blue}{25.08} (30.24)    
                        
& 41.19 & 52.65 & 31.74  & 33.81 & \textbf{89.27}\\
F1-score $\uparrow$         & \textcolor{blue}{21.69} (28.04) & \textcolor{blue}{21.25} (27.96)
                            & \textcolor{blue}{32.94} (38.57) & \textcolor{blue}{37.83}  (42.08)      

& 53.47 & 69.84 & 43.63 & 44.17 & \textbf{95.24}\\
\midrule

\multicolumn{10}{l}{\textbf{Held-out in-house testset}} \\
\midrule
IOU $\uparrow$          & \textcolor{blue}{5.23} (9.12) & \textcolor{blue}{5.46} (9.45)
                        & \textcolor{blue}{8.28} (14.00) & \textcolor{blue}{10.10} (17.66)                       
                        
                        & 20.12 & 32.08 & 35.62 & 27.39    &\textbf{72.09}\\
                        
DSC $\uparrow$              & \textcolor{blue}{9.58} (14.33) & \textcolor{blue}{9.69} (14.80)
                            & \textcolor{blue}{13.59} (18.04) & \textcolor{blue}{15.88} (24.57)                      
                            
                            & 27.80 & 38.63 & 41.55 & 34.26    &\textbf{83.78}\\

F1-score $\uparrow$         & \textcolor{blue}{13.03} (17.15) & \textcolor{blue}{13.17} (17.72)
                            & \textcolor{blue}{19.07} (25.26) & \textcolor{blue}{22.49} (30.07)

& 39.42 & 50.76 & 54.97 & 45.88    &\textbf{91.76}\\
\bottomrule
\end{tabular}
}
\caption{Comparison of segmentation performance on both in-domain and out-of-domain benchmarks.
LISA~\cite{lai2024lisa}, LISA++~\cite{yang2023lisa++}, VisionReasoner~\cite{liu2025visionreasoner}, and SAM4MLLM~\cite{chen2024sam4mllm} are re-implemented using their official codebases. We evaluate two settings: (1) directly loading their publicly released model weights (shown in \textcolor{blue}{blue}); and (2) further fine-tuning these models on our SFT and RL training datasets, which include the same images, masks, and QA pairs used for IBISAgent (shown in ``()''), ensuring fair comparison.
MedPLIB~\cite{huang2025towards}, Citrus-V~\cite{wang2025citrusvadvancingmedicalfoundation}, MMedAgent~\cite{li2024mmedagent}, and UniBiomed~\cite{wu2025unibiomed} are also re-implemented following their official repositories. For these models, we directly load the released weights without additional fine-tuning, as they were pretrained on large-scale public datasets that partially overlap with ours.}
\vspace{-10pt}
\label{tab:seg}
\end{table*}

\subsection{Agentic Reinforcement Learning}
\label{sec3.3}
We further optimize IBISAgent through RL with carefully designed rewards, enabling it to adaptively discover new action strategies and achieve higher-level decision-making, thereby moving beyond the constraints of mimicking the action trajectories learned during SFT.

\noindent \textbf{Dataset Curation.}
Unlike $\mathcal{D}_{\text{cold}}$, the RL training data includes only images, GT masks, and QA pairs, without click trajectories or reasoning trace annotations. This design encourages the model to autonomously explore and strengthen its pixel-level reasoning ability during RL. 

Specifically, we randomly sample image–mask pairs from the BiomedParseData~\citep{zhao2024biomedparse}. Following the same QA generation procedure as in the SFT stage, we obtain $564$K VQA instances. In addition, we incorporate widely used biomedical VQA datasets that do not require fine-grained pixel-level reasoning or segmentation. This hybrid data composition enables IBISAgent to selectively activate pixel-space reasoning only when necessary. In total, $888$K VQA samples are used for RL training, forming the dataset $\mathcal{D}_{\text{rl}}$. More details of $\mathcal{D}_{\text{rl}}$ are in the Appendix~\ref{sec:RLData}. 

\noindent \textbf{Reward Design.}
Unlike prior works~\cite{liu2025seg, huang2025sam, yan2025medreasoner, liu2025visionreasoner} that rely on overly simplified, outcome-only reward designs, we introduce a novel, fine-grained, rule-based reward framework that delivers dense feedback throughout the reasoning process. This enables the model to develop more efficient and effective decision-making strategies.
Formally, the reward framework consists of the following components:

\noindent\textbullet~ Format reward $\mathcal{S}_{\text{format}}$, which evaluates the structural validity of the model’s output $R$. It checks whether all required special tokens appear in the correct order and whether the \texttt{<action>} fields can be successfully parsed according to the predefined schema.

\noindent\textbullet~ Final-answer reward $\mathcal{S}_{ans}$,
which encompasses multiple task types. For close-ended QA questions, we simply check the exact match between the predicted and answers. For segmentation task, we compute the IoU between the predicted masks and GT masks and assign piecewise rewards based on predefined IoU thresholds.

\noindent\textbullet~ \textbf{Region-based click placement reward} $\mathcal{S}_{\text{click}}$ is defined as a bonus granted only when the model produces a reasonable click action. Specifically, given the model-predicted click $a_t$, we use the segmentation tool to generate the corresponding mask $M_t$ and compute the false-positive (FP) and false-negative (FN) regions between $M_t$ and the GT mask $M_{\text{gt}}$. A positive click is expected to fall within an FN region, while a negative click should lie within an FP region. Rewards and penalties are assigned accordingly, encouraging the model to place clicks in semantically meaningful locations rather than arbitrarily.

\noindent\textbullet~ \textbf{Progressive segmentation improvement reward} $\mathcal{S}_{\text{pseg}}$.
This reward enforces that each action $a_t$ leads to a segmentation improvement over the previous step. In other words, the mask produced after executing $a_t$ must achieve a higher quality than the mask at step $t-1$. This mechanism encourages the agent to consistently refine the segmentation rather than performing redundant actions or oscillating among repetitive operations. Concretely, we compute the IoU of the generated mask at each step $t$; if the score surpasses that of the mask from the previous action $a_{t-1}$, the agent receives a reward; otherwise, no reward is given.

\noindent\textbullet~ \textbf{Trajectory length reward} $\mathcal{S}_{\text{len}}$.
If the action sequence to complete segmentation is shorter than a predefined threshold, a reward is given; otherwise, penalties increasing with trajectory length are applied to encourage efficiency.

The final reward $\mathcal{S}$ is derived as: $\mathcal{S} = \frac{1}{5}( \mathcal{S}_{ans} + \mathcal{S}_{format} + \mathcal{S}_{\text{click}} + \mathcal{S}_{\text{pseg}} + \mathcal{S}_{\text{len}})$. Formal equations for each reward component are provided in Appendix~\ref{sec:rewardformula}.
Our fine-grained reward scheme better reflects the complexity of iterative segmentation, guiding the model to produce actions that are both spatially valid and semantically accurate.


\noindent \textbf{Optimization.}
Based on the rollout formulation and rewards defined above, we optimize the policy using GRPO~\citep{guo2025deepseek} without the KL penalty term~\citep{hu2025open} on dataset $\mathcal{D}_{rl}$:
\begin{equation}
\resizebox{0.9\linewidth}{!}{$
\begin{aligned}
\mathcal{L}_{\mathrm{RL}}
=
\mathbb{E}_{\substack{(I,Q,A)\sim \mathcal{D}_{\mathrm{rl}}\\ \{P_i\}_{i=1}^{G}\sim \pi_{\theta_{\mathrm{old}}}(\cdot \mid I,Q)}}
&\Big(
-\frac{1}{G}\sum_{i=1}^{G}\frac{1}{N_i}\sum_{t=1}^{T_i}
\min\!\big(\pi_{\theta_{i,t}} \mathcal{A}_i,\; \\
& \operatorname{clip}(\pi_{\theta_{i,t}},\,1-\epsilon,\,1+\epsilon)\,\mathcal{A}_i\big)
\Big),
\end{aligned}
$}
\end{equation}
\begin{equation}
\pi_{\theta_{i,t}}
=
\frac{\pi_{\theta}\left(r_{i,t},a_{i,t}\mid I,Q,P_{i,<t}\right)}
     {\pi_{\theta_{\mathrm{old}}}\left(r_{i,t},a_{i,t}\mid I,Q,P_{i,<t}\right)}.
\end{equation}
Here, 
$G$ is the number of rollout reasoning paths; $P_i=\{(r_{i,t},a_{i,t},o_{i,t})\}_{t=1}^{T_i}$ denotes the $i$-th reasoning path; $N_i$ is the total length of $P_i$ excluding observation tokens; $\mathcal{S}_i$ is the reward of $P_i$; and $\pi_{\theta}$ and $\pi_{\theta_{\mathrm{old}}}$ represent the current and old policy distributions, respectively. The normalized score $\mathcal{A}_i
=[\mathcal{S}_i - \mathrm{mean}(\{\mathcal{S}_j\}_{j=1}^{G})]/{\mathrm{std}(\{\mathcal{S}_j\}_{j=1}^{G})},$ reflects the relative quality of each reasoning path within the rollout group, enabling the model to distinguish between learnable and poor reasoning trajectories. Through RL training, the agent learns to adaptively reason over pixel features and plan click actions when necessary, achieving superior autonomous interactive segmentation.

\section{Experiment}

\subsection{Experimental Setup}
\textbf{Evaluation Benchmarks.}
For segmentation performance evaluation, we conduct experiments on three datasets:
(1) In-domain test set $D_{\text{test}}$. We randomly sample $9$K image–mask pairs from the official BiomedParseData~\cite{zhao2024biomedparse} test split. 
(2) Out-of-domain benchmark. We adopt the MeCOVQA-G+~\cite{huang2025multimodallargelanguagemodel, wang2025citrusvadvancingmedicalfoundation} test set, which includes 3K samples across 5 imaging modalities. This benchmark pairs biomedical images with text queries that explicitly request segmentation of specific anatomical structures or lesions. (3) Held-out in-house dataset. Since many foundation models and MLLMs are trained on large-scale public datasets, training–testing overlap may vary across prior works.
To ensure fair evaluation and avoid unintentional data leakage, we additionally evaluate on a private held-out dataset collected from three medical centers, comprising 1K CT, MRI, and pathology images across seven cancer types. For VQA performance evaluation, we use four widely adopted medical VQA benchmarks: PathVQA~\citep{he2020pathvqa}, SLAKE~\citep{liu2021slake}, VQA-RAD~\citep{lau2018dataset}, and OmniMedVQA~\citep{hu2024omnimedvqa}. 
More benchmark details are provided in Appendix~\ref{sec:testbench}.

\noindent \textbf{Metrics.}
For the segmentation task, we report mIOU and Dice score for mask segmentation, and F1 Score for mask-to-entity correspondence accuracy. For the VQA task, we evaluate the accuracy of the model’s responses.

\noindent \textbf{Implementation Details.}
We implement IBISAgent based on Qwen2.5-VL-7B~\citep{qwen2-5-vl}. MedSAM2~\cite{ma2025medsam2} is used as the interactive segmentation tool.
The training is conducted on a cluster of 16 NVIDIA A100 GPUs. For the cold-start SFT stage, we optimize the model with a learning rate of $1 \times 10^{-5}$ for $10$ epochs. The total batch size is $256$. The subsequent RL optimization is implemented using the VERL~\citep{sheng2025hybridflow} framework, where we set the training batch size to $256$ and generate $4$ candidate reasoning paths per question, up to a maximum of $20$ times of actions. We use a constant learning rate of $1 \times 10^{-6}$ and set the maximum context length to $32$K tokens. RL training runs for $12$ epochs. 


\subsection{Comparison with Previous SOTA Methods}
\textbf{Segmentation Performance.} We compare IBISAgent against two groups of baselines in~\cref{tab:seg}:
(1) General-purpose MLLMs with segmentation capability, including LISA-7B~\cite{lai2024lisa}, LISA++ 7B~\cite{yang2023lisa++}, SAM4MLLM-8B~\cite{chen2024sam4mllm}, and VisionReasoner-7B~\cite{liu2025visionreasoner} (using SAM2~\cite{ravi2024sam}). 
(2) Medical MLLMs that support segmentation, including MedPLIB-7B~\cite{huang2025towards}, Citrus-V 8B~\cite{wang2025citrusvadvancingmedicalfoundation}, UniBiomed-1B~\cite{wu2025unibiomed}, and MMedAgent-7B~\cite{li2024mmedagent} (using MedSAM~\cite{Ma_2024}).

Compared with general-domain MLLMs, IBISAgent achieves substantially higher performance across all benchmarks. Since these models were trained solely on datasets containing natural images, we further fine-tune them using our cold-start SFT and RL datasets—which include the same images, masks, and QA pairs used for training IBISAgent—to ensure fairness. Even after this additional fine-tuning, IBISAgent still significantly outperforms these models, demonstrating that its superiority does not simply arise from the use of a specialized dataset. Instead, the improvements primarily stem from our novel and effective design, which emulates the annotation process of human experts, enabling multi-round reasoning and refinement. This domain-specific modification, tailored for medical imaging, ultimately yields markedly better segmentation results. 

Compared with existing medical-specific MLLMs, IBISAgent achieves substantially superior segmentation performance. On average, it surpasses these models by at least $35.13\%$ in IoU, $37.58\%$ in DSC, and $29.79\%$ in F1 score. Notably, Citrus-V~\cite{wang2025citrusvadvancingmedicalfoundation} and UniBiomed~\cite{wu2025unibiomed} were trained on datasets far larger than ours; nevertheless, our method consistently outperforms them, further confirming that the improvements arise from the effectiveness of our proposed technical components rather than from the use of a specialized dataset. In particular, MMedAgent~\cite{li2024mmedagent} is also a tool-augmented MLLM that employs MedSAM~\cite{Ma_2024} for segmentation. IBISAgent still markedly outperforms this model, demonstrating that its advantages do not merely arise from the integration of segmentation tools, but from more accurate grounding and iterative reasoning-based segmentation mask refinement.

\begin{figure}
    \centering
    \includegraphics[width=\linewidth]{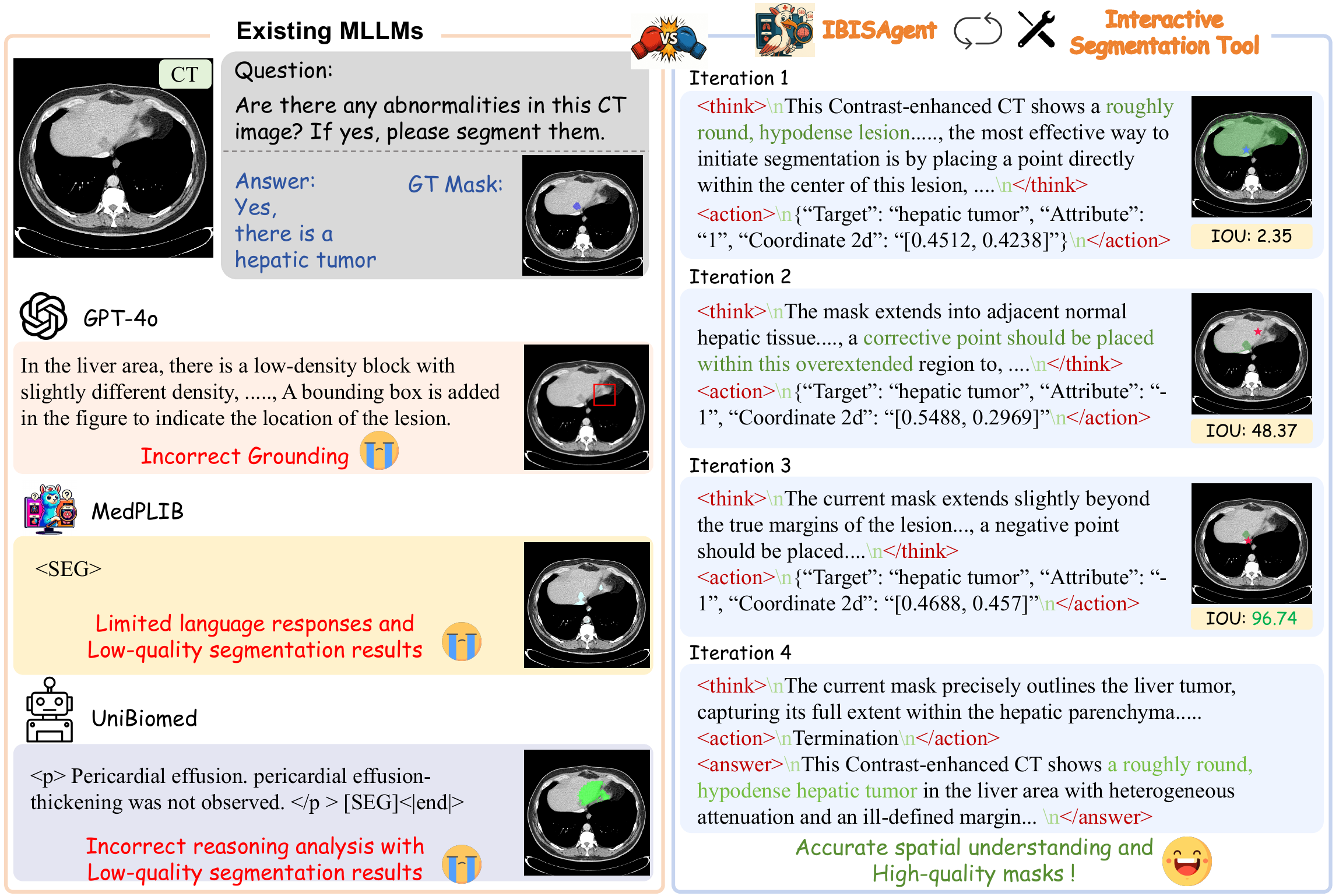}
    
    \caption{
        \textbf{Qualitative Analysis.} We present the responses and segmentation outputs on a reasoning–segmentation example. Existing MLLMs exhibit incorrect reasoning and low-quality segmentation, highlighting their misaligned fine-grained vision–language understanding. In contrast, \textbf{IBISAgent} delivers substantially improved reasoning quality and segmentation performance.
    }
    \label{fig:comparison}
\end{figure}

\begin{table}
  \Huge
  \centering
  \renewcommand{\arraystretch}{1.15}
  \resizebox{\linewidth}{!}
  {
  \begin{tabular}{@{}c|c|c|c|c|c@{}}
    \toprule
    \diagbox{Efficiency}{Method} & UniBiomed~\cite{wu2025unibiomed}  & MedPLIB~\cite{huang2025towards} & Citrus-V~\cite{wang2025citrusvadvancingmedicalfoundation} & MMedAgent~\cite{li2024mmedagent} & IBISAgent\\
    \hline
     Inference Time (s)       & 5.82 & 3.14 & 8.25  & 10.42  & 28.70   \\
    \midrule
  \end{tabular}
  }
\caption{Computational efficiency comparison. All experiments are conducted on the same A100 GPU.}
  \vspace{-10pt}
  \label{tab:Efficiency}
\end{table}

\begin{table}[t]
\huge
\centering
\renewcommand{\arraystretch}{1.15}
\resizebox{\linewidth}{!}{
\begin{tabular}{l|ccc|ccc}
\toprule
\multirow{2}{*}{Methods} & \multicolumn{3}{c|}{MeCOVQA-G+}  & \multicolumn{3}{c}{In-House Testset}\\
\cline{2-4} \cline{5-7}
& IoU $\uparrow$ & DSC $\uparrow$ & F1 $\uparrow$ & IoU $\uparrow$ & DSC $\uparrow$ & F1 $\uparrow$\\
\hline
\rowcolor{gray!20} GPT-4o~\cite{hurst2024gpt} + MedSAM2~\cite{ma2025medsam2}                   &  11.75 & 17.39 & 22.42  &  7.23 & 10.40 & 15.16\\
LLaVA-Med~\cite{li2023llava} + MedSAM2~\cite{ma2025medsam2}                 &  24.54 & 31.38 & 35.70   & 20.03 & 26.94 & 37.75\\
HuatuoGPT-Vision~\cite{chen2024huatuogpt} + MedSAM2~\cite{ma2025medsam2}    &  35.86 & 43.41 & 54.79   & 30.25 & 36.72 & 52.28\\
Lingshu~\cite{lingshu} + MedSAM2~\cite{ma2025medsam2}                       &  39.63 & 47.18 & 60.31   & 31.19 & 37.55 & 54.02\\
Chiron~\cite{sun2025enhancing} + MedSAM2~\cite{ma2025medsam2}               &  39.58 & 47.16 & 60.25   & 30.75 & 36.90 & 53.37\\
\hline 
\rowcolor{red!6} IBISAgent &  \textbf{80.63} & \textbf{89.27} & \textbf{95.24}   &  \textbf{72.09} & \textbf{83.78} & \textbf{91.76} \\ 
\bottomrule
\end{tabular}
}
\caption{Comparison of IBISAgent with tool-augmented MLLM agents. Except for the closed-source models, all competing methods in this table are implemented using their official open-source code and model weights, followed by further fine-tuning on our $\mathcal{D}_{\text{cold}}$ and $\mathcal{D}_{\text{rl}}$ datasets to ensure a fair comparison.}
\vspace{-10pt}
\label{tab:ab1}
\end{table}

\noindent \textbf{Qualitative Analysis.} \label{qualitative}
As shown in~\cref{fig:comparison}, we present qualitative results that further demonstrate the merits of IBISAgent. 
For a case requiring reasoning to identify a liver tumor, the powerful closed-source GPT-4o~\cite{hurst2024gpt} outputs seemingly correct textual reasoning but produces an incorrect localization bounding box. This indicates that the model fails to align textual reasoning with visual features—it may be hallucinating plausible answers rather than truly understanding the medical image, ultimately leading to incorrect localization. MedPLIB~\cite{huang2025towards} provides coarse localization that roughly captures the lesion region but suffers from false-positive predictions. Moreover, its language output is highly constrained—when performing segmentation, the model rigidly outputs only the token “\texttt{<SEG>}”, losing the rich language generation capability originally inherent to large language models. UniBiomed~\cite{wu2025unibiomed} produces incorrect textual reasoning and responses, describing objects that do not exist in the image and generating low-quality segmentation masks. This observation reinforces our motivation: current methods that rely on implicit ``\texttt{<SEG>}'' tokens for segmentation disrupt the MLLM’s native text output space, compromising both its language capability and semantic generalization. 
Moreover, these approaches exhibit fundamentally limited pixel-level visual reasoning, preventing the model from truly understanding fine-grained visual features. In contrast, IBISAgent not only generates correct and coherent textual reasoning but also adaptively produces high-quality masks through step-by-step, precise pixel grounding. This capability arises from our design that decouples reasoning from segmentation, thereby preserving the MLLM’s inherent language reasoning ability while simultaneously enabling adaptive, multi-round refinement.

\noindent \textbf{Efficiency Comparison.}
We randomly sampled $1,000$ cases ($30\%$) from the out-of-domain testset MeCOVQA-G+~\cite{huang2025multimodallargelanguagemodel, wang2025citrusvadvancingmedicalfoundation} to measure the average time required by different models to process each sample. The results are reported in~\cref{tab:Efficiency}. Because IBISAgent performs multi-round mask refinement, its per-case inference time is longer than that of existing MLLMs. Fundamentally, such multi-step interactions between the agent and the environment are intended to trade additional computation time for improved performance—an inherent limitation of multi-round agent systems. Nevertheless, the inference overhead of IBISAgent remains within an acceptable range, especially in light of the substantial performance gains it delivers.

\subsection{Ablation Studies}
\textbf{IBISAgent vs. Prompting-based MLLM Agents.} 
To validate the effectiveness of formulating segmentation as an interleaved fine-grained visual reasoning and action planning Markov  decision process, we design several baseline agent workflows that directly predict bounding boxes or points using MLLMs, followed by segmentation with MedSAM2~\cite{ma2025medsam2}. We then compare IBISAgent with these agent systems to evaluate the advantages of our iterative, multi-round, reasoning-driven formulation. 
Specifically, we construct different agent systems using GPT-4o~\cite{hurst2024gpt}, LLaVA-Med~\cite{li2023llava}, HuatuoGPT-Vision~\cite{chen2024huatuogpt}, Lingshu~\cite{lingshu}, and Chiron~\cite{sun2025enhancing}. The results are shown in~\cref{tab:ab1}. IBISAgent consistently outperforms all competing agent systems, demonstrating that our method effectively activates the model’s intrinsic pixel-level reasoning ability, enabling advanced segmentation performance that cannot be achieved through simple workflow-style tool calling.

\noindent \textbf{Effectiveness of Training Strategies.}
To evaluate the effectiveness of our training framework, we compare IBISAgent with several Qwen2.5-VL-7B–based variants: (1) $\mathcal{M}_{base}$, a prompt-driven baseline equipped with MedSAM2~\cite{ma2025medsam2}; (2) $\mathcal{M}_{cold}$, trained solely with cold-start SFT; (3) $\mathcal{M}_{cold+reflect}$, trained with cold-start SFT augmented by synthesized self-reflective trajectories; (4) $\mathcal{M}_{rl}$, trained exclusively with RL; and (5) $\mathcal{M}_{cold+rl}$. The results in Table~\ref{tab:ablation_training_stages} demonstrate the effectiveness of our training strategies. The prompt-only approach is neither adaptable nor robust, whereas progressively incorporating our proposed training stages yields substantial improvements over the base model, highlighting the necessity of each component. The RL stage provides the largest performance gain, indicating that RL’s exploration–exploitation dynamics and reward feedback are crucial for acquiring context-aware decision-making policies and enabling genuine vision-centric multi-step reasoning. Furthermore, by integrating all training strategies, IBISAgent surpasses all baselines, further supporting our motivation for developing a versatile training framework that strengthens the model’s pixel-level visual reasoning capabilities.
\begin{table}[t]
\centering
\renewcommand{\arraystretch}{1.15}
\resizebox{\linewidth}{!}{
\begin{tabular}{lccccccc}
\toprule
\multirow{2}{*}{\textbf{Method}} &
\multicolumn{3}{c}{\textbf{Training Strategies}} &
\multicolumn{3}{c}{\textbf{In-House Testset}} \\
\cmidrule(lr){2-4}\cmidrule(lr){5-7}
& \textbf{SFT} & \textbf{Self-Reflection} & \textbf{RL}
& IoU $\uparrow$ & DSC $\uparrow$ & F1 $\uparrow$  \\
\midrule
\rowcolor{gray!20} \textit{prompt-driven} $\mathcal{M}_{\text{base}}$ &  &  &           & 11.77 & 16.83 & 23.47  \\
\addlinespace[0.3em]\midrule
$\mathcal{M}_{\text{cold}}$                    & \checkmark &            &              & 53.42 & 62.01 & 68.61 \\
$\mathcal{M}_{\text{cold+reflect}}$            & \checkmark & \checkmark &              & 57.16 & 67.73 & 74.52 \\
$\mathcal{M}_{\text{rl}}$                      &  &                      & \checkmark   & 62.77 & 71.29 & 77.50 \\
$\mathcal{M}_{\text{cold+rl}}$                 & \checkmark &  &  \checkmark            & 68.92 & 78.08 & 85.44 \\
\rowcolor{red!6} \textbf{IBISAgent}            & \checkmark & \checkmark & \checkmark   & \textbf{72.09} & \textbf{83.78} & \textbf{91.76} \\
\bottomrule
\end{tabular}
}
\caption{Ablation study on our training strategies. Checkmarks (\checkmark) indicate which strategies are applied.}
\label{tab:ablation_training_stages}
\vspace{-10pt}
\end{table}

\begin{table}[t]
\centering
\huge
\renewcommand{\arraystretch}{1.15}
\resizebox{\linewidth}{!}{
\begin{tabular}{ccc|cccc|cccc}
\toprule
\multicolumn{3}{c|}{\textbf{Reward Signals}} &
\multicolumn{4}{c|}{\textbf{MeCOVQA-G+}} & \multicolumn{4}{c}{\textbf{In-House Testset}} \\
\cline{1-3}\cline{4-7}\cline{8-11}
 \textbf{$\mathcal{S}_{\text{click}}$} & \textbf{$\mathcal{S}_{\text{pseg}}$} & \textbf{$\mathcal{S}_{\text{len}}$}
& IoU $\uparrow$ & DSC $\uparrow$ & F1 $\uparrow$ & $T_{avg}$ $\downarrow$  & IoU $\uparrow$ & DSC $\uparrow$ & F1 $\uparrow$ & $T_{avg}$ $\downarrow$\\
\hline
\rowcolor{gray!20}              &       &                       & 73.77 & 82.62 & 88.53 & 11.29  & 68.96 & 79.17 & 87.06 & 13.44\\
\hline
                     \checkmark &            &                  & 76.60 & 85.77 & 91.23 & 10.61    & 70.45 & 81.23 & 89.30 & 12.74\\
                                 & \checkmark &                 & 77.64 & 86.85 & 92.31 & 8.59    & 70.62 & 81.40 & 89.56 & 10.07\\
                                  &            & \checkmark     & 74.19 & 82.88 & 89.05 & 5.94    & 69.03 & 79.65 & 87.48 & 7.22\\
                \checkmark & \checkmark &                       & 80.61 & 89.19 & \textbf{95.24} & 8.12          & 72.05 & 83.74 & 91.71 & 9.68\\
                 \checkmark &            & \checkmark           & 77.73 & 86.88 & 92.34 & 5.03             & 70.75 & 81.63 & 89.78 & 6.49\\
               &       \checkmark     & \checkmark              & 79.37 & 88.25 & 93.97 & 4.26             & 71.16 & 82.28 & 90.55 & 5.43\\
\rowcolor{red!6}              \checkmark & \checkmark & \checkmark   &  \textbf{80.63} & \textbf{89.27} & \textbf{95.24} & \textbf{3.67}            
                                            & \textbf{72.09} & \textbf{83.78} & \textbf{91.76} & \textbf{4.89}\\
\bottomrule
\end{tabular}
}
\caption{Ablation study on reward design. The gray row indicates the baseline that excludes all segmentation-tailored rewards and uses only the standard format and answer rewards.}
\label{tab:ablation_rewards}
\vspace{-10pt}
\end{table}
\noindent \textbf{Significance of Different Reward Signals.}
We demonstrate the effectiveness of the proposed Region-based Click Placement Reward $\mathcal{S}_{\text{click}}$, Progressive Segmentation Improvement Reward $\mathcal{S}_{\text{pseg}}$, and Trajectory Length Reward $\mathcal{S}_{\text{len}}$ using the ablation results in \cref{tab:ablation_rewards}. We begin by removing all three reward signals to establish a baseline that relies solely on the standard format reward and final answer reward (shown in the gray row). We then progressively incorporate different combinations of our proposed reward signals to examine their impact on segmentation performance. We also report the average predicted trajectory length (steps) to comprehensively evaluate how these rewards influence both segmentation quality and interaction efficiency. Each reward contributes substantially to the overall performance. The rewards $\mathcal{S}_{\text{click}}$ and $\mathcal{S}_{\text{pseg}}$ provide the greatest improvements in mask quality, as they ensure accurate click localization and encourage each interaction step to make a positive contribution to the segmentation outcome. Meanwhile, $\mathcal{S}_{\text{pseg}}$ and $\mathcal{S}_{\text{len}}$ improve interaction efficiency by discouraging redundant or uninformative clicks and preventing unnecessarily long trajectories, enabling the model to learn when to stop. By integrating all reward components, IBISAgent achieves an optimal balance between segmentation performance and interaction efficiency.
\section{Conclusion}
In this paper, we present IBISAgent, a novel MLLM-based agent capable of pixel-level visual reasoning for unified biomedical object referring and segmentation. IBISAgent reformulates segmentation as a multi-step Markov Decision Process and enhances MLLMs’ pixel-level understanding without introducing additional model components. We develop IBISAgent through a two-stage training protocol and design fine-grained rewards that incentivize continuous self-improvement reasoning. Additionally, we introduce a large-scale high-quality dataset with thinking and action trajectories. Through comprehensive empirical evaluation, we demonstrate the competitive performance of IBISAgent across diverse segmentation tasks. Our study paves the way for future exploration of vision-centric, multi-step decision-making agents for holistic medical image analysis.
\par

\section*{Acknowledgments}
This work is supported by the National Natural Science Foundation of China under Grant No. 62502429, and the Zhejiang Key Laboratory Project (2024E10001).

\par
{
    \small
    \bibliographystyle{ieeenat_fullname}
    \bibliography{main}
}

\clearpage
\setcounter{page}{1}
\maketitlesupplementary

\appendix 

\section{Dataset Details}
\subsection{Dataset for SFT}

Our dataset $\mathcal{D}_{cold}$ for cold-start Supervised Fine-Tuning (SFT) is a large-scale, high-quality collection of textual reasoning and action trajectories for interactive segmentation. It comprises a total of $47,146$ individual samples (e.g., slices or images), which collectively contain $456,795$ visual question-answer (VQA) pairs that capture a wide range of diverse scenarios.
The dataset is meticulously curated by filtering segmentation trajectories to ensure high fidelity, achieving an overall average Intersection over Union (IoU) of $94.27$ (median: $95.07$) and an average Dice score of $0.9703$ (median: $97.47$). This high level of accuracy confirms the quality of the segmentation ground truths generated by the trajectories. Furthermore, the average trajectory length per sample is $8.69$ steps, indicating a rich capture of the multi-step refinement processes required for complex segmentation tasks.

\subsubsection{Modality Diversity}

The dataset is characterized by its extensive diversity, covering $9$ distinct medical imaging modalities. This broad range ensures that models trained on this data can generalize across various imaging types, from common modalities like CT and MRI to more specialized ones like Pathology and OCT. A detailed breakdown of the dataset composition by modality is presented in \cref{tab:sft_modality_stats}.

\begin{table*}[h!]
  \centering
  \caption{Statistical overview of the \textbf{SFT} dataset $\mathcal{D}_{cold}$, categorized by imaging modality. Detailed breakdown by task group is shown in \cref{tab:sft_task_stats}.}
  \label{tab:sft_modality_stats}
  \resizebox{\textwidth}{!}{%
  \begin{tabular}{lccccccc}
    \toprule
    \textbf{Modality} & \textbf{Samples} & \textbf{Total QAs} & \textbf{Avg. Length} & \textbf{Avg. IoU} & \textbf{Median IoU} & \textbf{Avg. DSC} & \textbf{Median DSC} \\
    \midrule
    \textbf{CT}              & 22,504 & 209,291 & 8.30 & 0.9462 & 0.9519 & 0.9722 & 0.9753 \\
    \textbf{Dermoscopy}      &  1,302 &  13,086 & 9.05 & 0.9348 & 0.9338 & 0.9662 & 0.9658 \\
    \textbf{Endoscope}       &  2,431 &  17,434 & 6.17 & 0.9548 & 0.9609 & 0.9768 & 0.9800 \\
    \textbf{Fundus}          &    472 &   4,850 & 9.28 & 0.8834 & 0.8772 & 0.9378 & 0.9346 \\
    \textbf{MRI}             &  7,453 &  82,616 & 10.08 & 0.9404 & 0.9457 & 0.9691 & 0.9721 \\
    \textbf{OCT}             &    205 &   1,850 & 8.02 & 0.9116 & 0.9074 & 0.9537 & 0.9515 \\
    \textbf{Pathology}       &  1,195 &  11,765 & 8.85 & 0.9081 & 0.9243 & 0.9501 & 0.9607 \\
    \textbf{Ultrasound}      &  3,499 &  36,972 & 9.57 & 0.9341 & 0.9360 & 0.9659 & 0.9669 \\
    \textbf{X-Ray}           &  8,085 &  78,931 & 8.76 & 0.9460 & 0.9542 & 0.9721 & 0.9766 \\
    \midrule
    \textbf{Total}      & \textbf{47,146} & \textbf{456,795} & \textbf{8.69} & \textbf{0.9427} & \textbf{0.9507} & \textbf{0.9703} & \textbf{0.9747} \\
    \bottomrule
  \end{tabular}
  }
\end{table*}

\subsubsection{Task Diversity}

In addition to modality diversity, the dataset spans $38$ distinct segmentation tasks. These tasks are derived from a combination of the Medical Segmentation Decathlon (MSD) dataset \cite{simpson2019largeannotatedmedicalimage}, covering both organs (e.g., `liver', `heart') and tumors (e.g., `brain tumor'), and various other specialized public biomedical datasets (e.g., `ACDC', `LIDC-IDRI', `GlaS') derived from BioMedParse \cite{zhao2024biomedparse}. This task diversity exposes the model to a wide array of anatomical structures, pathologies, and image characteristics. The detailed statistics for each task group are provided in \cref{tab:sft_task_stats}.

\subsection{Dataset for RL}
\label{sec:RLData}
\subsubsection{Pixel-Level Reasoning Corpus for RL.}
In addition to the cold-start SFT dataset, we curated a large-scale, high-quality corpus specifically for the RL stage. As detailed in \cref{sec3.3}, this corpus is also sampled from BioMedParse~\cite{zhao2024biomedparse} including some samples used in $\mathcal{D}_{cold}$. It contains only the images, ground-truth masks, and QA pairs that require identifying fine-grained visual cues and localizing the specified mask region, thereby compelling the agent to autonomously explore and learn effective action policies.

This RL corpus comprises a total of $60,826$ samples, which collectively contain $564,385$ QA pairs. 
Similar to the SFT dataset, this corpus is highly diverse, spanning $8$ medical imaging modalities and $39$ distinct segmentation tasks. Detailed statistical breakdowns by modality and task group are provided in \cref{fig:rl_modal} and \cref{fig:rl_dataset_detail}, respectively.

\subsubsection{Commonly-Used Medical VQA Datasets.}
During RL training, we jointly use common medical VQA datasets and our curated pixel-level reasoning corpus. This hybrid training strategy preserves the model’s visual understanding and question-answering abilities, enabling the policy to selectively invoke pixel-space reasoning only when necessary. \cref{tab:VQANum} reports the number of VQA pairs in the medical VQA datasets used in our training. 



\subsection{Testing Benchmarks}
\label{sec:testbench}
\subsubsection{In-domain test set \texorpdfstring{$\mathcal{D}_{test}$}{D\_test}}
To comprehensively evaluate the model's generalization capabilities and robustness in pixel-level reasoning, we constructed a diverse test set $\mathcal{D}_{test}$ that is strictly disjoint from the training corpora ($\mathcal{D}_{cold}$ and the $\mathcal{D}_{rl}$ corpus). This dataset is designed to assess performance across a wide spectrum of medical imaging modalities and challenging segmentation scenarios.

The test set comprises a total of $9,902$ samples, containing $156,289$ VQA pairs. 
Unlike the training sets, which emphasize breadth by covering a wide variety of tasks, the test set focuses on more complex and challenging scenarios. Specifically, the test set places greater emphasis on fine-grained visual feature recognition—such as identifying tiny structures, intricate abnormalities, lesions, and tumors—to ensure that the benchmark rigorously evaluates the model’s ability to handle challenging targets. Consequently, the test set spans $32$ distinct task groups.

\begin{figure*}[t]
    \centering
    \begin{subfigure}[b]{0.49\textwidth}
        \centering
        \includegraphics[width=\linewidth]{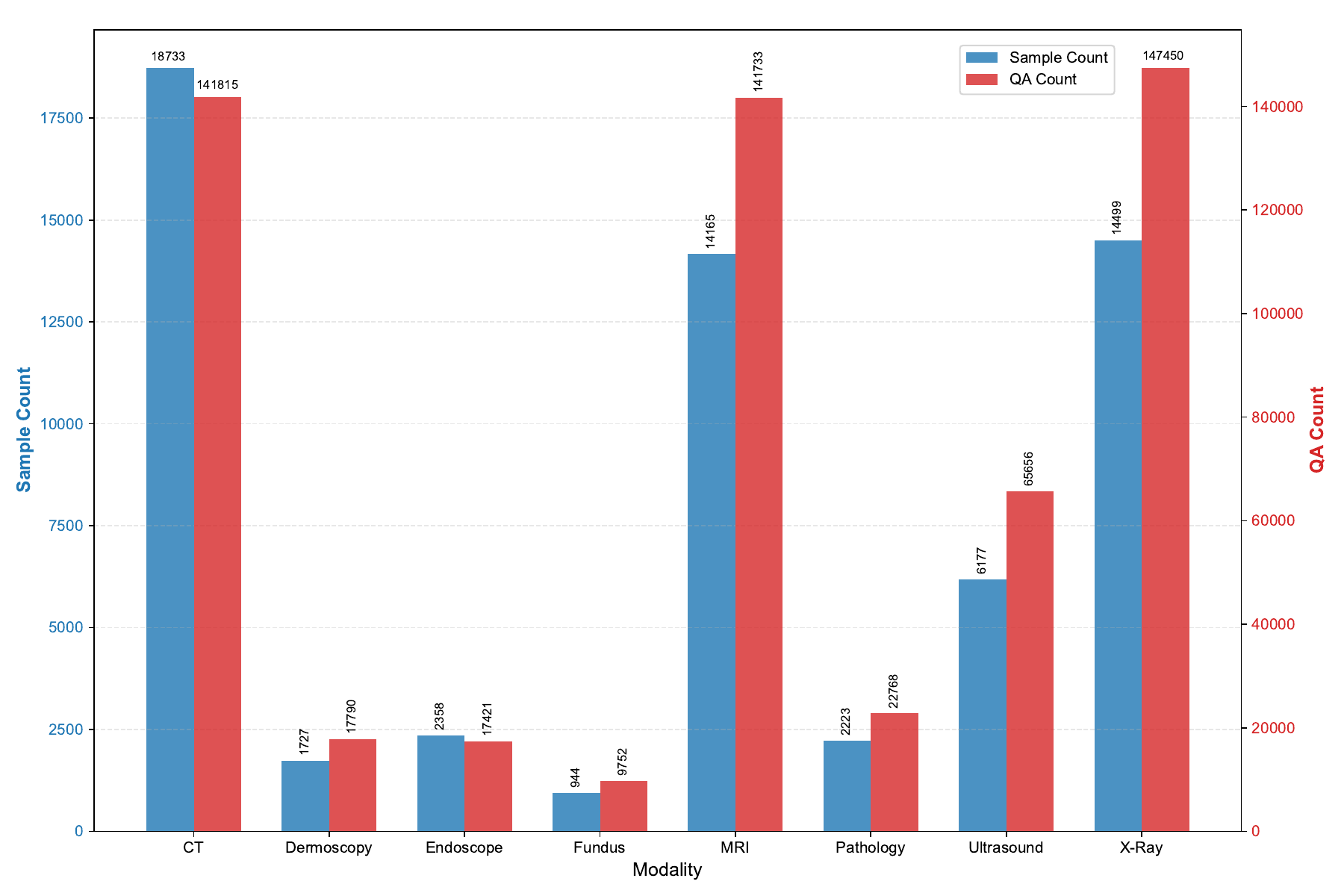}
        \caption{\textbf{RL Corpus $\mathcal{D}_{rl}$:} Statistics by Modality}
        \label{fig:rl_modal}
    \end{subfigure}
    \hfill 
    \begin{subfigure}[b]{0.49\textwidth}
        \centering
        \includegraphics[width=\linewidth]{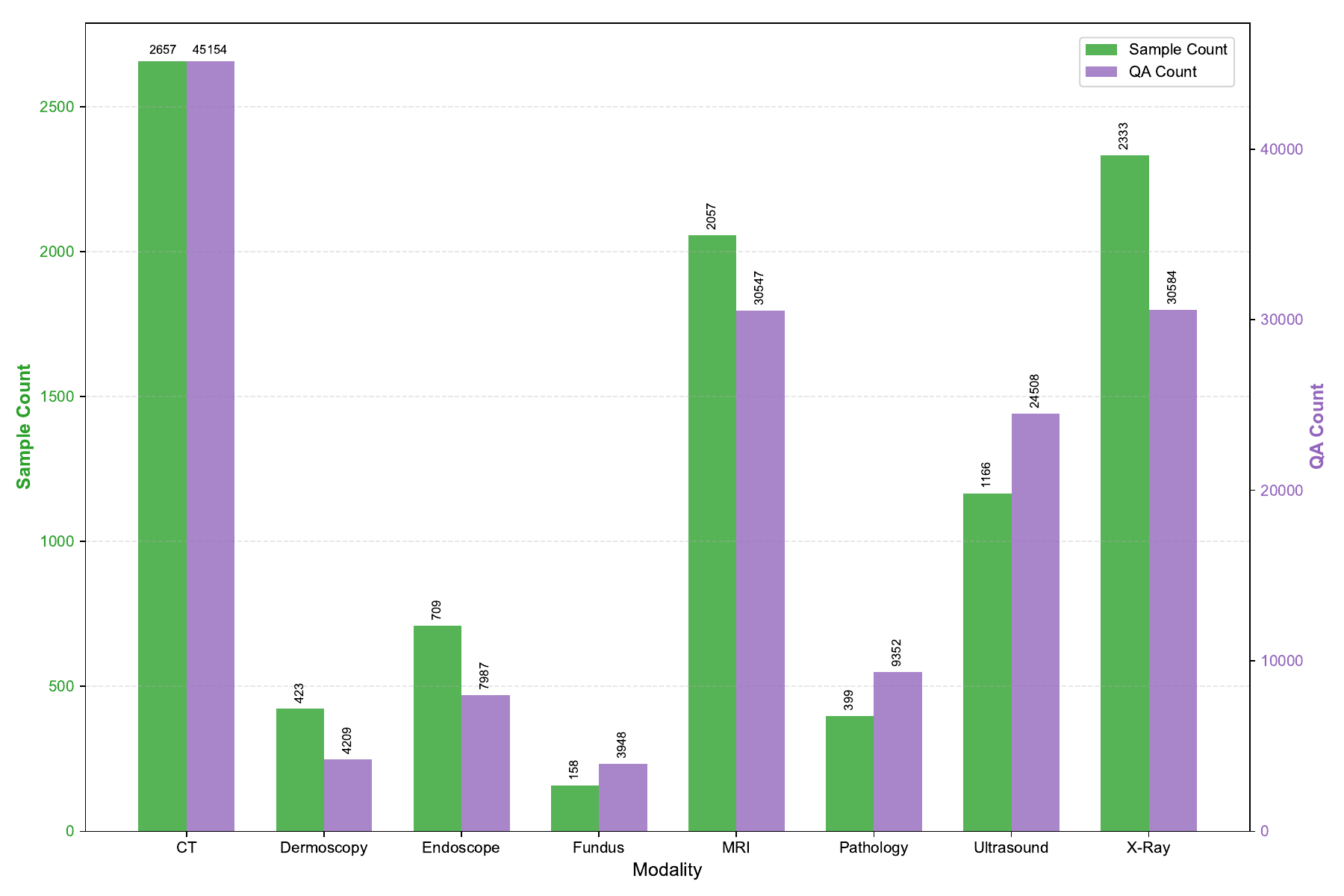}
        \caption{\textbf{In-domain Test Set $\mathcal{D}_{test}$:} Statistics by Modality}
        \label{fig:test_modal}
    \end{subfigure}
    
    \caption{\textbf{Modality distribution.} \textbf{(a)} The RL corpus $\mathcal{D}_{rl}$ (training stage) contains $60,826$ samples and $564,385$ QAs across $8$ modalities. \textbf{(b)} The In-domain Test set $\mathcal{D}_{test}$ comprises $9,902$ samples and $156,289$ QAs covering the same 8 modalities. The dual-axis plots show the sample count (left axis) and total QA pairs (right axis) for each category.}
    \label{fig:modality_stats}
\end{figure*}

\begin{figure*}[h!]
    \centering
    
    \begin{subfigure}[b]{0.99\textwidth}
        \centering
        \includegraphics[width=\linewidth]{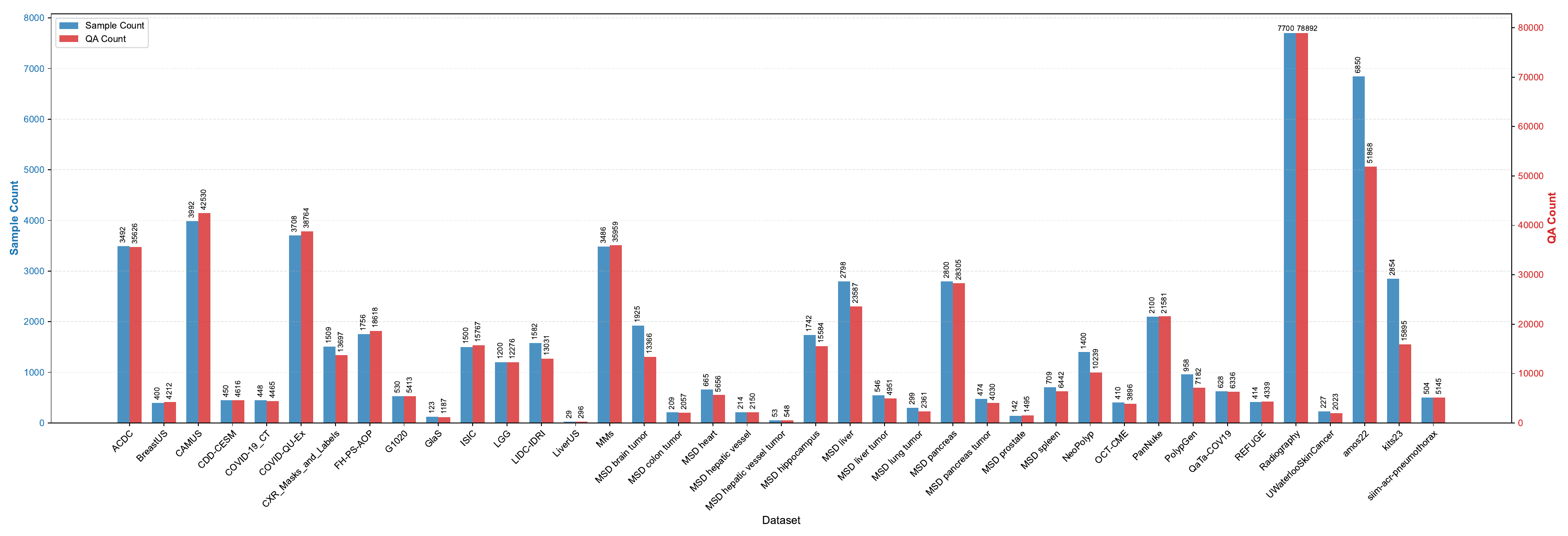}
        \caption{\textbf{RL Corpus $\mathcal{D}_{rl}$:} Detailed statistics across 39 task groups (datasets).}
        \label{fig:rl_dataset_detail}
    \end{subfigure}
    
    \vspace{0.5cm} 
    
    \begin{subfigure}[b]{0.99\textwidth}
        \centering
        \includegraphics[width=\linewidth]{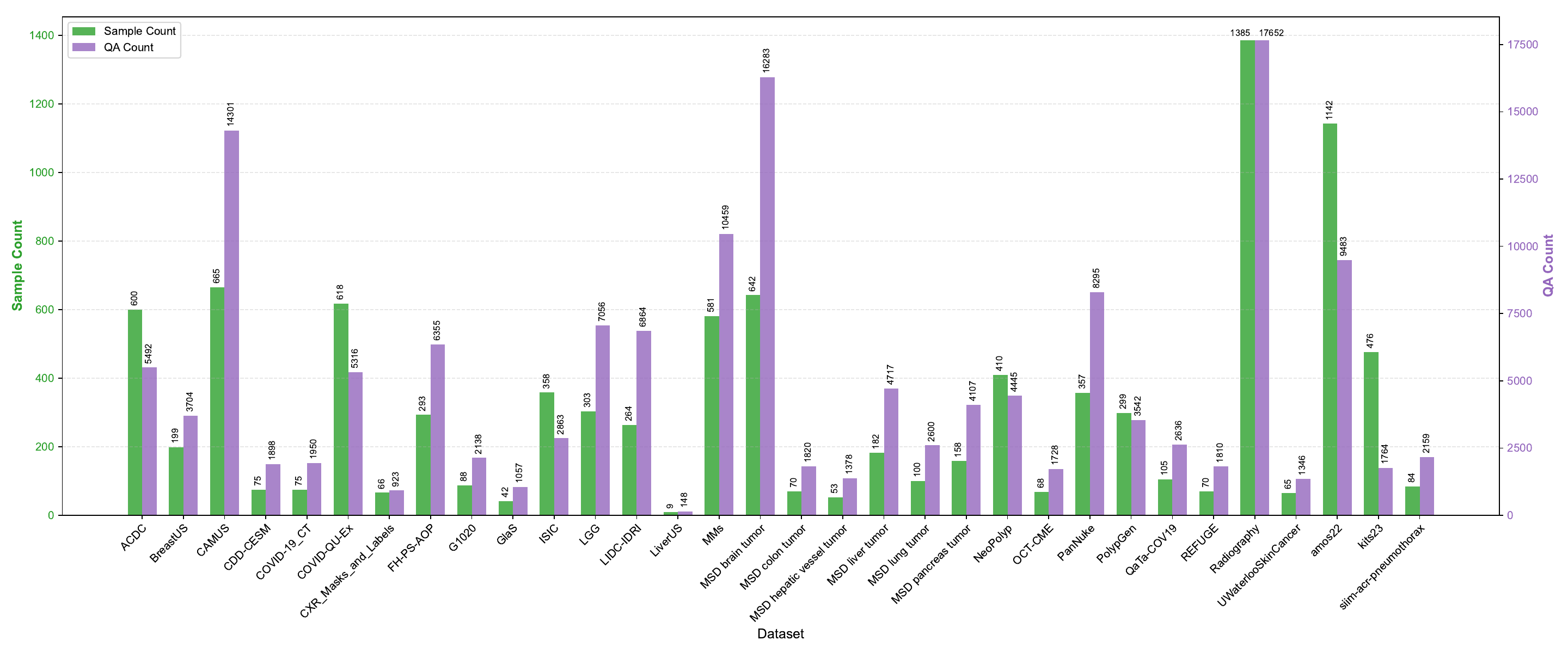}
        \caption{\textbf{In-domain Test Set $\mathcal{D}_{test}$:} Detailed statistics across 32 task groups (datasets).}
        \label{fig:test_dataset_detail}
    \end{subfigure}
    
    \caption{\textbf{Detailed breakdown by sub-dataset.} This figure illustrates the diversity of our data. \textbf{(a)} The RL corpus $\mathcal{D}_{rl}$ includes 39 distinct datasets. \textbf{(b)} The In-domain Test set $\mathcal{D}_{test}$ covers 32 datasets. Note the dual-axis scale for Sample Count (left) and QA Count (right).}
    \label{fig:dataset_detailed_stats}
\end{figure*}

In terms of modality, $\mathcal{D}_{test}$ covers $8$ major medical imaging types: CT, MRI, X-Ray, Ultrasound, Pathology, Endoscopy, Dermoscopy, and Fundus. 
This distribution ensures that the evaluation reflects real-world clinical diversity. 
The statistical distributions of samples and QA pairs by modality and by specific dataset are visualized in \cref{fig:test_modal} and \cref{fig:test_dataset_detail}, respectively.

\begin{table}[ht]
\centering
\caption{The distribution of commonly-used medical VQA datasets used in RL stage. ``HuatuoV\_A" and ``HuatuoV\_I" refer to the Huatuo\_PubMedVision\_Alignment and Huatuo\_PubMedVision\_InstructionTuning VQA datasets, respectively.}
\label{tab:VQANum}
\renewcommand{\arraystretch}{1.1} 
\resizebox{0.5\linewidth}{!}
{\begin{tabular}{lc}
\hline
\textbf{Dataset}     & \textbf{VQA Number} \\ \hline
\textbf{HuatuoV\_A}             & 128000           \\
\textbf{HuatuoV\_I}             & 128000           \\
\textbf{PMC\_VQA}                & 32000           \\
\textbf{VQA\_RAD}                 & 8000             \\
\textbf{SLAKE}                    & 9000             \\
\textbf{PATH\_VQA}                & 19000            \\ \hline
\end{tabular}}
\end{table}

\subsubsection{Out-of-domain test set MeCOVQA-G+}
Datasets for training and evaluating text-segmentation alignment in the medical domain are extremely scarce. One of the few publicly available resources is MeCoVQA-G, which was recently introduced alongside the MedPlib paper~\cite{huang2025towards}. MeCoVQA-G is a large-scale, pixel-level VQA subset of the MeCoVQA family, constructed by pairing biomedical images with natural-language questions that explicitly ask the model to segment a given anatomical structure or lesion. Each sample contains a 2D image slice, a templated question targeting a specific anatomical class, and the corresponding binary segmentation mask as the ground-truth answer. The released split is $100$K training pairs and $2,344$ test pairs.

MeCOVQA-G+~\cite{wang2025citrusvadvancingmedicalfoundation} is a thoroughly re-annotated and expanded edition of the MeCoVQA-G~\cite{huang2025towards}. MeCOVQA-G+ increases both the scale and modality diversity of its predecessor, delivering a more reliable and comprehensive benchmark for medical text-to-segmentation tasks. MeCOVQA-G+ comprises $3,157$ carefully curated text–segmentation pairs. The samples span a wide range of modalities, including X-ray, CT, MRI, ultrasound, and endoscopy. Each image has been meticulously reviewed
by a team of medical experts to ensure the accuracy of the segmentation masks.

\subsubsection{Held-out in-house test set}
For testing, in addition to our in-domain test set $\mathcal{D}_{\text{test}}$, we use a completely held-out in-house dataset comprising 1k CT, MRI, and histopathology images across $7$ cancer types for zero-shot evaluation. In this held-out set, CT images include 100 liver tumor cases, 100 gallbladder tumor cases, 100 pancreatic cancer cases, and 100 kidney tumor cases. MRI images include 100 colon tumor cases and 100 brain cancer cases. Histopathology images include 400 breast cancer cases. For each case, human annotators construct a VQA pair along with a corresponding reasoning trajectory.

\section{More Implementation Details}
\subsection{Trajectory Generation}
\label{sec:TrajGen}
To train our model for multi-step medical image segmentation, we require a dataset of expert-like interaction trajectories. We employed an automated algorithm to generate these trajectories by simulating the sequential refinement process an expert annotator would perform, leveraging the click simulation strategy proposed in \cite{xu2016deep}. The algorithm iteratively interacts with a pre-trained interactive segmentation model (specifically MedSAM2 \cite{ma2025medsam2}), intelligently placing clicks to correct errors in the current prediction until it closely matches the ground truth.

The core of this method is a deterministic, greedy strategy for selecting the next interaction point, augmented by a \textbf{mask prompting mechanism} to ensure stability. At each step $t$, we strictly utilize the low-resolution mask logits from the previous step, denoted as $M_{logits}^{(t-1)}$, alongside the cumulative click history $H_t$. This simulates a realistic annotation workflow where the annotator refines an existing mask rather than starting from scratch at each interaction.

First, we identify the error regions in the current prediction $M_{p}^{(t)}$. The False Negative (FN) region, $M_{fn}$, represents the target area missed by the model, while the False Positive (FP) region, $M_{fp}$, represents areas incorrectly included in the prediction:
\begin{equation}
    M_{fn} = M_{gt} \setminus M_{p}^{(t)}, \quad M_{fp} = M_{p}^{(t)} \setminus M_{gt}
\end{equation}

To emulate human behavior prioritizing large error regions, we compute the Euclidean distance transform for both error masks, denoted as $D_{fn}$ and $D_{fp}$. The algorithm selects the next click $a_t = (c_t, l_t)$ by targeting the pixel with the maximum distance value (i.e., the center of the largest error region):
\begin{equation}
    (c_t, l_t) = 
    \begin{cases} 
        (\arg\max D_{fn}, 1) & \text{if } \max(D_{fn}) \geqslant \max(D_{fp}) \\
        (\arg\max D_{fp}, 0) & \text{otherwise}.
    \end{cases}
\end{equation}

\begin{figure*}[htbp]
    \centering
    \includegraphics[width=\textwidth]{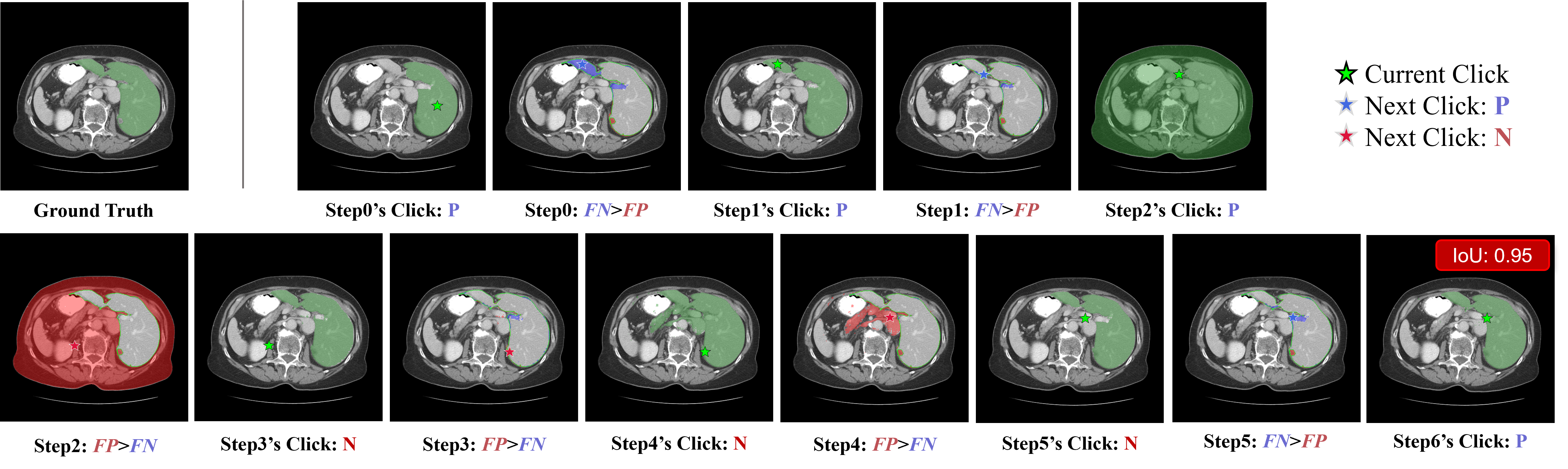}
    \caption{\textbf{An illustrative example of the automated trajectory generation process for liver segmentation.} The algorithm progressively refines the predicted mask through iterative interactions. 
    \\ For each iteration (e.g., Step 0), two visualization panels are presented: 
    (1) The \textbf{first image} displays the \textit{current segmentation state}, showing the predicted mask (green translucent overlay) generated by the current click (marked by a green star). 
    (2) The \textbf{second image} illustrates the \textit{error analysis} against the Ground Truth (delineated by a \textbf{\textcolor[HTML]{00FF00}{green outline}}). The differences are visualized as blue translucent regions for False Negatives (FN, under-segmentation) and red translucent regions for False Positives (FP, over-segmentation). The star in this panel indicates the calculated \textbf{next action} based on the largest error region: a \textbf{\textcolor{blue}{blue star}} denotes a Positive Click (P) to correct under-segmentation, while a \textbf{\textcolor{red}{red star}} denotes a Negative Click (N) to correct over-segmentation.}
    \label{fig:sample_image}
\end{figure*}

Crucially, the update rule for the segmentation model $\mathcal{S}$ incorporates both the updated history and the dense mask prompt from the previous iteration:
\begin{equation}
    M_{p}^{(t+1)}, M_{logits}^{(t+1)} = \mathcal{S}(I, H_{t} \cup \{a_t\}, M_{logits}^{(t)})
\end{equation}
where $M_{logits}^{(0)}$ is initialized as None. This iterative process continues until the IoU exceeds a threshold $\tau_{iou}$ or the maximum step count $T_{max}$ is reached. The implementation logic is detailed in \cref{alg:trajectory_generation_python}. An example process is depicted as \cref{fig:sample_image}.

\begin{algorithm}[htbp]
\caption{Pseudo-code for Trajectory Generation}
\label{alg:trajectory_generation_python}
\begin{lstlisting}[language=Python]
def generate_trajectory(image, gt_mask, model, max_steps=20, iou_thresh=0.95):
    pred_mask = np.zeros_like(gt_mask)
    low_res_logits = None
    history = []

    for step in range(max_steps):
        # Check Stop Condition
        iou = calculate_iou(pred_mask, gt_mask)
        if iou >= iou_thresh:
            break

        # Identify Error Regions (FN & FP) and Find Click
        fn_mask = (gt_mask == 1) & (pred_mask == 0)
        fp_mask = (gt_mask == 0) & (pred_mask == 1)
        
        fn_dist = cv2.distanceTransform(fn_mask.astype(np.uint8), ...)
        fp_dist = cv2.distanceTransform(fp_mask.astype(np.uint8), ...)

        if np.max(fn_dist) >= np.max(fp_dist):
            coords = np.unravel_index(np.argmax(fn_dist), fn_dist.shape)
            history.append((coords, 1))
        else:
            coords = np.unravel_index(np.argmax(fp_dist), fp_dist.shape)
            history.append((coords, 0))
        
        # Update Prediction using Points AND Previous Mask Logits
        points = [p[0] for p in history]
        labels = [p[1] for p in history]
        
        pred_masks, scores, logits = model.predict(
            point_coords=points,
            point_labels=labels,
            mask_input=low_res_logits
        )

        # Select best mask and update logits for next iteration
        best_idx = np.argmax(scores)
        pred_mask = pred_masks[best_idx]
        low_res_logits = logits[best_idx] 

    return history, pred_mask
\end{lstlisting}
\end{algorithm}

\subsection{QA Generation}
\label{sec:QAGen}

To train IBISAgent to understand diverse user intents and perform pixel-level tasks, we constructed a comprehensive instruction dataset. We utilized Gemini-2.5-Pro to generate a rich set of visual question-answering (VQA) pairs and instructions. 

\noindent\textbf{Hallucination Prevention via Fact-Based Generation.} A critical challenge in generating medical instructions is preventing the LLM from hallucinating non-existent anatomical features or pathologies. To mitigate this, we strictly conditioned the generation process on ground-truth evidence. Specifically, we provided Gemini-2.5-Pro with the raw image, the ground-truth mask, and a verified caption of the biomedical object. The model was explicitly instructed to generate prompts \textbf{only} based on these visible facts, ensuring that every instruction (e.g., ``Segment the left lung'') corresponds to an object actually present in the image.

\begin{figure*}[t]
    \centering
    \includegraphics[width=\linewidth]{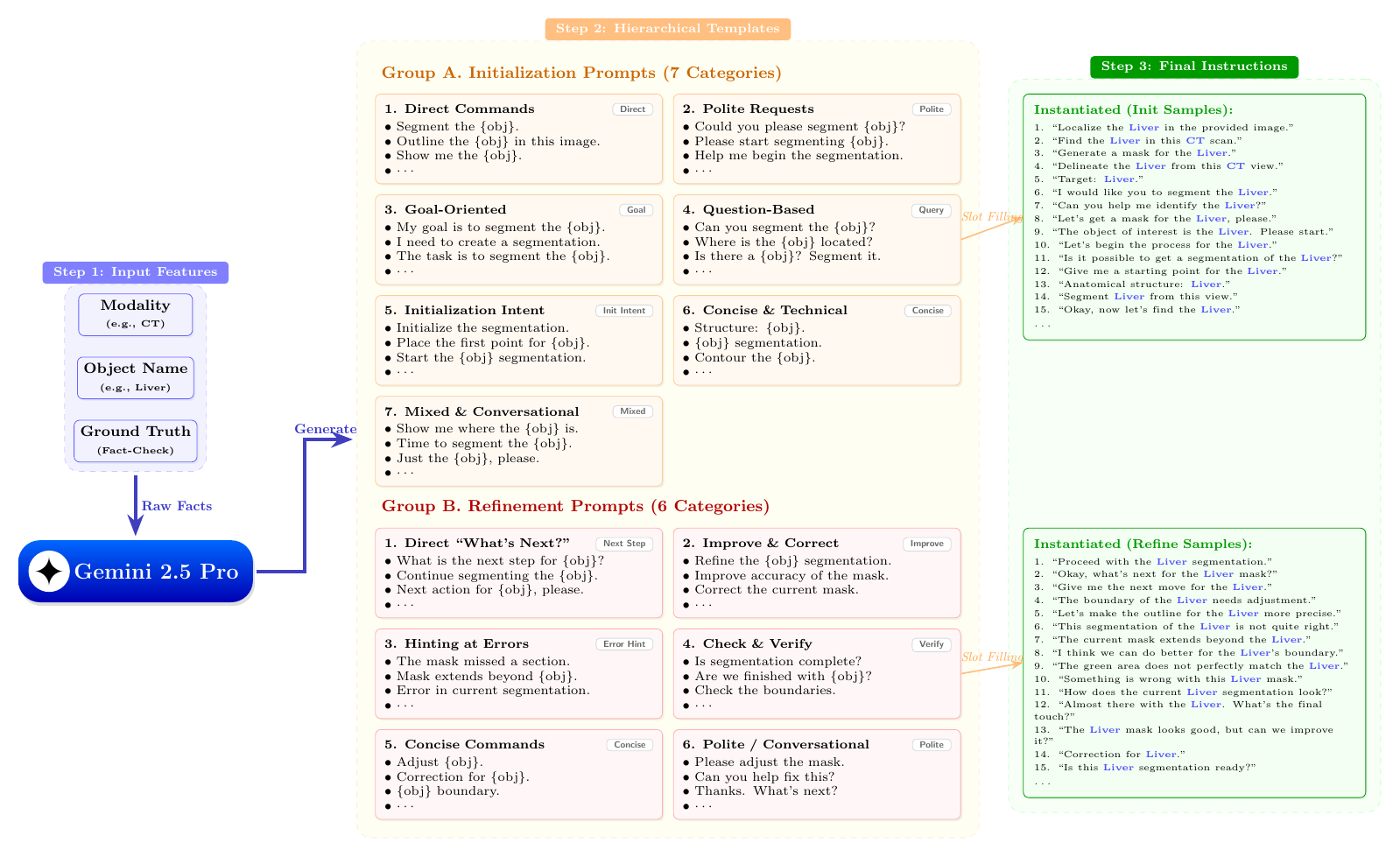}
    \caption{\textbf{The hierarchical prompt generation pipeline.} 
    To ensure both diversity and factual accuracy, we leverage \textbf{Gemini 2.5 Pro} to synthesize a comprehensive instruction library. 
    Conditioned on ground-truth input features (Step 1), the model dynamically generates a taxonomical prompt set (Step 2) divided into \textbf{Group A: Initialization Prompts} ($7$ categories, covering imperative to conversational tones) and \textbf{Group B: Refinement Prompts} ($6$ categories, focusing on iterative corrections). 
    These templates are then instantiated into final instructions (Step 3), creating a rich dataset ($50$ items per group) that covers diverse user intents while strictly adhering to visual facts.}
    \label{fig:b2}
\end{figure*}

\noindent\textbf{Hierarchical Instruction Categories.} We designed a taxonomical prompt library to cover different phases of the segmentation process, as illustrated in \cref{fig:b2}:

\begin{itemize}
    \item \textbf{Initialization Prompts.} These prompts initiate the segmentation task from scratch. To mimic real-world user behavior, we categorized them into $7$ broad types, ranging from \textit{Direct Commands} to \textit{Goal-Oriented} statements. 
    
    \textbf{Specialized Query Templates.} We applied a $70/30$ split between standard imperative prompts and \textbf{Interrogative Queries} to enhance the model's flexibility. For the latter, we designed $5$ specific sub-templates to mimic clinical uncertainty:
    \begin{enumerate}[label=(\roman*)]
        \item \textit{Conditional Logic:} ``Is there a \{object\_name\}? If so, please segment it.''
        \item \textit{Compound Requests:} ``Can you find and segment the \{object\_name\}?''
        \item \textit{Clinical Protocol Tone:} ``I need to verify the presence of a \{object\_name\}...''
        \item \textit{Indirect/Conversational:} ``I'm wondering if there's a \{object\_name\}...''
        \item \textit{Concise Checks:} ``Visible \{object\_name\}? Please provide segmentation.''
    \end{enumerate}
    
    \item \textbf{Refinement Prompts.} We categorized them into $6$ types. These prompts are used during the iterative interaction steps. They focus on fine-grained adjustments, such as \textit{Requesting Next Steps} (e.g., ``What is the next step?''), \textit{Error Correction} (e.g., ``The mask extends beyond the boundary''), and \textit{Verification} (e.g., ``Is this segmentation complete?'').
\end{itemize}

\noindent\textbf{Diverse Assistant Response Generation.} To ensure that the agent's output is naturalistic and varied rather than robotic, we also constructed a template library for the \textbf{Assistant's final responses}. These are categorized into $5$ styles: \textit{Direct \& Concise} (``Segmentation complete.''), \textit{Confident Affirmation} (``The object has been successfully segmented.''), \textit{Object-Referencing} (``The \{object\_name\} is fully segmented.''), \textit{Question-Answering} (``Yes, the object was found...''), and \textit{Conversational} (``All done!''). This diversity prevents the model from overfitting to a single termination phrase.

During data construction, we dynamically fill all user and assistant templates with the specific anatomical target name (e.g., ``left ventricle'') and imaging modality (e.g., ``MRI''), ensuring high relevance and grammatical correctness.

\subsection{Reasoning Generation for Our SFT Dataset}
\label{sec:ReasonGen}

A cornerstone of our SFT dataset is the high-quality, step-by-step reasoning (\texttt{<think>...</think>}) that accompanies each agent action. Generating this data presents a significant challenge: our agent (M1) operates from a limited, first-person perspective (seeing only the current segmentation mask), but the optimal reasoning for its next action (e.g., ``correcting an over-segmentation'') requires an ``oracle'' or ``ground-truth'' perspective (knowing the precise False Positive and False Negative regions).

To solve this, we employed a ``Teacher-Student'' (or ``Oracle-Agent'') generation pipeline. We utilized the powerful \textbf{GPT-5} model as the ``Teacher'' (M2) to synthesize reasoning traces for our ``Student'' agent (M1).


Our core innovation lies in a sophisticated prompt strategy that leverages the advanced role-playing capabilities of GPT-5 to bridge the information gap between the Oracle and the Agent.

1. \textbf{Persona and Perspective Simulation.} The system prompt instructs M2 to adopt the persona of an ``expert radiologist.'' Critically, it commands M2 to generate reasoning strictly from the limited perspective of the junior agent. The prompt explicitly states: ``\textit{This agent ONLY SEES ONE THING: a single, combined green mask... Write as if you are genuinely deducing the next step from only the visible image.}'' This forces the teacher model to reverse-engineer the logic: instead of simply stating the error (which it knows), it must explain \textbf{why} the visual features (e.g., texture differences, anatomical landmarks) suggest an error exists.

\begin{figure}[h!]
    \centering 
    \subfloat[A Negative Click example for Prostate segmentation. The M2 (oracle) view (Left) shows the False Positive (FP, red) region. The M1 (agent) view (Right) only sees the combined green mask.]{
        \includegraphics[width=0.98\linewidth]{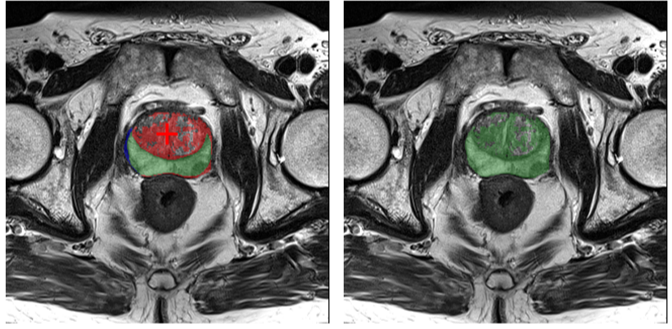}
        \label{fig:sub_neg_click}
    }
    \\
    \subfloat[A Positive Click example for Prostate segmentation. The M2 (oracle) view (Left) shows the False Negative (FN, blue) region. The M1 (agent) view (Right) only sees the incomplete green mask.]{
        \includegraphics[width=0.98\linewidth]{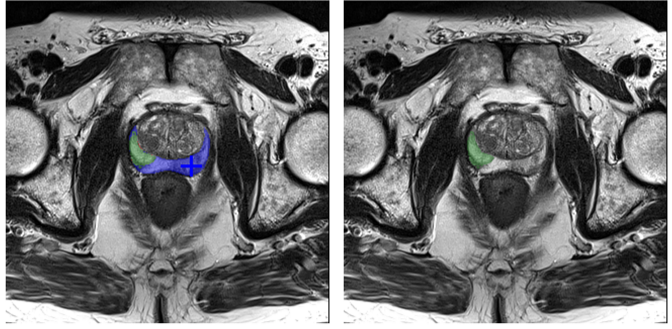}
        \label{fig:sub_pos_click}
    }
    \caption{
    \textbf{Comparison of the ``Oracle'' view (for M2) and the ``Agent'' view (for M1) used in SFT reasoning generation.}\\
    Notably, \textbf{this \textcolor[HTML]{608A60}{agent-visible mask (green)} is the sum of the oracle's \textcolor[HTML]{608A60}{True Positive (green)} and \textcolor[HTML]{FF0000}{False Positive (red)} areas.} M1 must learn to infer the expert's corrective reasoning from this limited perspective.
    }
    \label{fig:m1_m2_comparison}
\end{figure}

2. \textbf{Privileged Information.} While M2 writes from the agent's perspective, it views a privileged ``oracle'' image. As shown in \cref{fig:m1_m2_comparison} (Left), this image explicitly visualizes segmentation errors: a \textbf{Green Mask} for True Positives, a \textbf{Red Mask} for False Positives (over-segmentation), and a \textbf{Blue Mask} for False Negatives (under-segmentation). It also indicates the ``correct'' next action (Positive/Negative point).

3. \textbf{Preventing Information Leakage.} A primary risk in this pipeline is ``prompt leakage,'' where the teacher accidentally reveals its privileged knowledge (e.g., mentioning ``the red mask''). Thanks to the superior instruction-following capability of GPT-5 compared to smaller models, we effectively mitigated this using a robust set of \textbf{Forbidden Concepts}. The system prompt strictly prohibits the output from containing terms like `Red', `Blue', `Cross', `TP', `FP', `FN', or `Ground Truth'. 



\begin{figure*}[t]
    \centering
    \includegraphics[width=\linewidth]{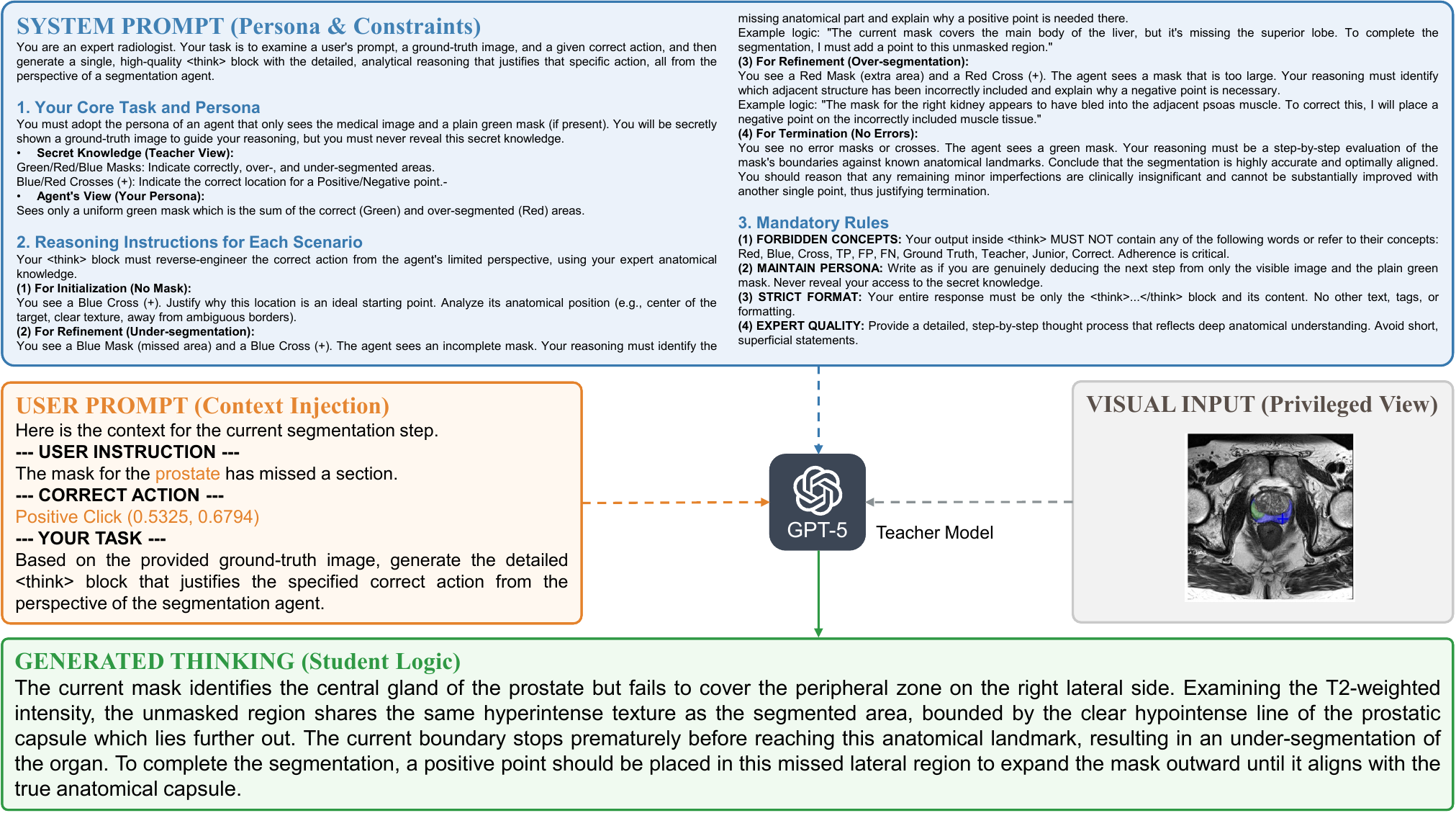}
    \caption{\textbf{The prompt engineering pipeline for synthesizing pixel-level reasoning traces.} To bridge the gap between ground-truth knowledge and the agent's limited perspective, we employ a Teacher-Student strategy using GPT-5. The pipeline integrates three key components: (1) a System Prompt that establishes an expert persona and enforces strict constraints; (2) the User Context containing the target instruction and correct action; and (3) the Privileged Oracle View where error regions are explicitly visualized. The model is tasked with ``reverse-engineering" the reasoning, producing vision-based justifications for the corrective action without revealing its access to the privileged information.}
    \label{fig:b3}
\end{figure*}

As shown in \cref{fig:b3}, this approach generates dense, anatomically grounded reasoning traces without requiring manual templates for every scenario (like we did before using smaller models).

\section{Reward Functions}
\label{sec:rewardformula}
\textbf{The reasoning-format reward.}
The reasoning-format reward $\mathcal{S}_{format}$ evaluates the structural validity of $R$ by verifying that the model’s output includes all required special tokens in the prescribed order. Specifically, the model should enclose its chain-of-thought between \texttt{<think>} and \texttt{</think>} tags, place the tool-call choices and parameters between \texttt{<tool\_call>} and \texttt{</tool\_call>} tags, and place the final answer between \texttt{<answer>} and \texttt{</answer>} tags. Outputs that adhere to this structure receive a positive reward.

\begin{equation}
\mathcal{S}_{\mathrm{format}} =
\begin{cases}
1, & \text{if all required fields appear} \\
   & \text{and are in the correct order}, \\
0, & \text{otherwise}.
\end{cases}
\end{equation}

\noindent \textbf{The final-answer reward.}
The final-answer reward $\mathcal{S}_{ans}$ encompasses multiple task types, thereby providing the agent with diverse feedback. For \textbf{multiple-choice questions}, we simply check the exact match between the predicted answer and the ground truth: 
\begin{equation}
S_{\mathrm{ans}}(A,\hat{A}) =
\mathbb{I}\!\left(A = \hat{A}\right).
\end{equation}
Here, $A$ denotes the ground-truth answer and $\hat{A}$ is the predicted answer obtained by rule-based parsing of the model’s final output.
The indicator $\mathbb{I}$ is defined to be $1$ if $A=\hat{A}$ and $0$ otherwise.

For \textbf{segmentation} tasks, in contrast to earlier reward designs, we use MedSAM2 as external reward providers. Given either click points predicted by the MLLM, we query MedSAM2 to obtain a segmentation mask. We then compute the IOU between this mask and the ground-truth mask and assign piecewise rewards as follows:
\begin{equation}
\mathcal{S}_{\text{ans}} =
\begin{cases}
3, & \mathrm{IoU} > 0.80,\\
2, & 0.70 < \mathrm{IoU} \leqslant 0.80,\\
1, & 0.50 < \mathrm{IoU} \leqslant 0.70,\\
0, & \text{otherwise}.
\end{cases}
\end{equation}
This segmentation reward supplies strong positive feedback only when the predicted region closely matches the ground truth, while at lower IoU levels it provides guidance that encourages gradual improvement. 

\noindent \textbf{Region-based click placement reward.}
A \textbf{core innovation} of our framework is that we introduce explicit spatial constraints into the reward signal. Unlike generic RL agents that may learn to click arbitrarily to trigger tool usage, we enforce semantically valid interactions through $\mathcal{S}_{\text{click}}$. This reward serves as a dense supervision signal, ensuring that the agent's actions are grounded in the anatomical reality of the image.

Let $a_t = (c_t, p_t)$ denote the action at step $t$, where $c_t \in \mathbb{R}^2$ is the spatial coordinate and $p_t \in \{+1, -1\}$ indicates a positive or negative click type. Let $M_{t-1}$ be the segmentation mask from the previous step (with $M_0$ initialized as an empty mask). We define the eligible error regions for interaction based on the ground truth mask $M_{gt}$:
\begin{align}
\Omega_{\text{FN}} = M_{gt} \setminus M_{t-1},\ \Omega_{\text{FP}} = M_{t-1} \setminus M_{gt}
\end{align}
A click is considered valid if a positive point falls within the under-segmented region ($\Omega_{\text{FN}}$) or a negative point falls within the over-segmented region ($\Omega_{\text{FP}}$). The reward function is formalized as:
\begin{equation}
\mathcal{S}_{\text{click}}(a_t) =
\begin{cases}
r_{\text{click}}, & \text{if } p_t = +1 \land c_t \in \Omega_{\text{FN}}, \\
r_{\text{click}}, & \text{if } p_t = -1 \land c_t \in \Omega_{\text{FP}}, \\
-\lambda_{\text{miss}}, & \text{otherwise}.
\end{cases}
\end{equation}
where $r_{\text{click}} = 1$ is a positive bonus for spatially accurate clicks, and $\lambda_{\text{miss}} = 1$ is a penalty for invalid clicks. This reward effectively guides the policy to minimize the symmetric difference between the predicted and ground-truth masks step-by-step.

\noindent \textbf{Progressive segmentation improvement reward.}
To prevent the agent from engaging in redundant operations or oscillating between states without improving the result, we incorporate a progressive improvement reward $\mathcal{S}_{\text{pseg}}$. This component evaluates the marginal contribution of each action to the overall segmentation quality.

Let $\text{IoU}(M, M_{gt})$ denote the Intersection-over-Union between a mask $M$ and the ground truth. We calculate the quality gain $\Delta \mathcal{Q}_t$ after executing action $a_t$:
\begin{equation}
\Delta \mathcal{Q}_t = \text{IoU}(M_t, M_{gt}) - \text{IoU}(M_{t-1}, M_{gt}).
\end{equation}
The reward is assigned only if the action yields a strictly positive gain:
\begin{equation}
\mathcal{S}_{\text{pseg}} = \mathbb{I}(\Delta \mathcal{Q}_t > 0)
\end{equation}
This incentivizes the agent to strictly ascend the gradient of segmentation quality.

\noindent \textbf{Trajectory length reward.}
Efficiency is a critical metric for clinical assistants. To encourage the model to achieve high-quality segmentation with the minimum number of interactions, we introduce a trajectory length reward $\mathcal{S}_{\text{len}}$.
Let $T$ be the total number of steps taken in a reasoning path, and $T_{\text{opt}}$ be a predefined optimal threshold. The reward is defined as:
\begin{equation}
\mathcal{S}_{\text{len}} =
\begin{cases}
r_{\text{eff}}, & \text{if } T \leqslant T_{\text{opt}}, \\
-\gamma \cdot (T - T_{\text{opt}}), & \text{if } T > T_{\text{opt}}.
\end{cases}
\end{equation}
where $r_{\text{eff}}=1$ is a bonus for efficient completion, and $\gamma=0.2$ is a decay factor that applies a linear penalty for each additional step beyond the threshold. This formulation balances the trade-off between exhaustive refinement and interaction efficiency.

\section{More Experiments}
\label{sec:extraexp}

\subsection{VQA Performance.} 
We also conducted experiments demonstrating that pixel-level reasoning not only improves segmentation performance but also enhances the model’s VQA capabilities. The results of evaluation across three medical VQA benchmarks are summarized in~\cref{tab:VQA}. IBISAgent outperforms both open-source and proprietary MLLMs. IBISAgent outperforms both open-source and proprietary MLLMs. Notably, compared with existing medical MLLMs trained on large-scale VQA datasets, IBISAgent achieves at least a \textbf{$2.4\%$} improvement in average accuracy. This further validates our motivation that enhancing MLLMs’ understanding of fine-grained medical image features fundamentally improves their medical image analysis capabilities. Pixel-level exploration of localized regions closely mirrors the way clinicians interpret and reason about medical images, and our work effectively stimulates and strengthens this critical capability in MLLMs.

\begin{table}[t]
\centering
\renewcommand{\arraystretch}{1.15}
\resizebox{\linewidth}{!}{
\begin{tabular}{l|ccc|c}
\toprule
Methods & VQA-RAD  & PathVQA & SLAKE & AVG.\\
\hline
GPT-4o~\cite{hurst2024gpt}    &  64.9 & 58.1 & 70.9  & 64.6 \\
LLaVA-Med-7B~\cite{li2023llava}  &  53.1 & 44.2 & 47.5 & 48.3 \\
HuatuoGPT-Vision-34B~\cite{chen2024huatuogpt}    &  62.0 & 51.3 & 69.5   & 60.9 \\
Lingshu~\cite{lingshu}        &  66.1 & 68.7 & 78.0   & 70.9 \\
Chiron~\cite{sun2025enhancing} & 72.7 & 68.9 & 77.3   & 73.0 \\
\hline 
\rowcolor{red!6} IBISAgent &  \textbf{73.4} & \textbf{69.2} & \textbf{83.5}   &  \textbf{75.4} \\ 
\bottomrule
\end{tabular}
}
\caption{Comparison of IBISAgent with existing MLLMs on different VQA benchmarks.}
\label{tab:VQA}
\end{table}

\subsection{Impact of Segmentation Tool Types}
We further examine the effect of replacing the interactive segmentation tool used by IBISAgent. Specifically, we substitute the default MedSAM~2~\cite{ma2025medsam2} with alternative tools, including MedSAM~\cite{Ma_2024}, SAM~\cite{kirillov2023segment}, and SAM~2~\cite{ravi2024sam2segmentimages}, and compare the resulting performance, as shown in~\cref{tab:ablation_tool_types}. We find that IBISAgent remains highly robust to the choice of interactive segmentation tool, exhibiting only minor differences in final segmentation accuracy across these replacements. This robustness arises from IBISAgent’s ability to generate precise click-point locations and perform multi-round refinement, which jointly help maintain segmentation quality and ensure strong test-time stability. These results indicate that IBISAgent can effectively adapt to a wide range of interactive segmentation tools, rather than relying solely on MedSAM~2.

\begin{table}[t]
\centering
\renewcommand{\arraystretch}{1.15}
\resizebox{\linewidth}{!}{
\begin{tabular}{lccccccc}
\toprule
\multirow{2}{*}{\textbf{Methods}} &
\multicolumn{3}{c}{\textbf{MeCOVQA-G+}} &
\multicolumn{3}{c}{\textbf{In-House Test set}} \\
\cmidrule(lr){2-4}\cmidrule(lr){5-7}
& IoU $\uparrow$ & DSC $\uparrow$ & F1 $\uparrow$
& IoU $\uparrow$ & DSC $\uparrow$ & F1 $\uparrow$  \\
\midrule
IBISAgent (SAM)& 79.95 &     88.74&     94.83  & 71.82 & 83.19 & 90.98 \\
IBISAgent (SAM 2)   & 80.32 & 89.01 &   95.08    & 71.93 & 83.40 & 91.36 \\
IBISAgent (MedSAM)    &  80.29 & 89.00  & 95.03   & 71.91 & 83.37 & 91.32 \\
\rowcolor{red!6} \textbf{IBISAgent (MedSAM 2)}            & \textbf{80.63} & \textbf{89.27} & \textbf{95.24}   & \textbf{72.09} & \textbf{83.78} & \textbf{91.76} \\
\bottomrule
\end{tabular}
}
\caption{Ablation study on segmentation tool types.}
\label{tab:ablation_tool_types}
\end{table}

\subsection{The Performance of the Segmentation Tools}
We also report the standalone performance of the segmentation tools on the test sets as a reference, further highlighting the superiority of our method. \cref{tab:toolperformance} presents the results. In this comparison, we evaluate segmentation performance both with and without GT bounding-box prompts. Because MedSAM and MedSAM 2 support only visual prompts, their results in the ``w/o bbox'' setting are marked as $\times$. From~\cref{tab:toolperformance}, we observe that IBISAgent consistently achieves the highest performance across both segmentation modes. These findings indicate that IBISAgent exhibits strong generalization in text-driven segmentation and, when using GT bbox as the first step, delivers segmentation quality that consistently surpasses the competing tools such as MedSAM, MedSAM 2, and BiomedParse.

Overall, these results further validate the advantages of IBISAgent. Under our formulation, the agent performs precise pixel-level visual reasoning to accurately localize target regions and iteratively refine masks, enabling segmentation performance that exceeds the inherent upper bound of the underlying segmentation tools themselves.

\begin{table*}[t]
\centering
\resizebox{\textwidth}{!}{
\begin{tabular}{l|cc|cc|cc|cc|cc|cc}
\toprule
\multirow{2}{*}{Models} &
\multicolumn{4}{c|}{In-domain test set} &
\multicolumn{4}{c|}{MeCOVQA-G+ } &
\multicolumn{4}{c}{In-House Test set} \\
\cline{2-13}
& \multicolumn{2}{c|}{w/o bbox} & \multicolumn{2}{c|}{with bbox} 
& \multicolumn{2}{c|}{w/o bbox} & \multicolumn{2}{c|}{with bbox}
& \multicolumn{2}{c|}{w/o bbox} & \multicolumn{2}{c}{with bbox} \\
& IOU & DSC & IOU & DSC
& IOU & DSC & IOU & DSC
& IOU & DSC & IOU & DSC \\ 
\midrule
SAM 2~\cite{Ma_2024} 
& $\times$ & $\times$ & 80.30 & 85.61 
& $\times$ & $\times$ & 65.12 & 75.54
& $\times$ & $\times$ & 51.08 & 56.65 \\

MedSAM~\cite{Ma_2024} 
& $\times$ & $\times$ & 79.73 & 85.44 
& $\times$ & $\times$ & 60.32 & 71.74
& $\times$ & $\times$ & 49.50 & 53.28 \\

MedSAM2~\cite{ma2025medsam2}
& $\times$ & $\times$ & 82.07 & 87.28 
& $\times$ & $\times$ & \underline{71.30} & \underline{81.12}
& $\times$ & $\times$ & \underline{59.24} & \underline{64.49} \\

BiomedParse~\citep{zhao2024biomedparse}
& 83.03 & 87.19 & \underline{84.28} & \underline{89.67} 
& 37.68 & 45.39 & 67.41 & 78.36
& 27.48 & 34.23 & 55.87 & 60.68 \\

\midrule

\textbf{IBISAgent}
& \textbf{85.58} & \textbf{92.21} & \textbf{86.37} & \textbf{92.48}
& \textbf{80.63} & \textbf{89.27} & \textbf{81.56} & \textbf{90.11}
& \textbf{72.09} & \textbf{83.78} & \textbf{72.96} & \textbf{84.83} \\

\bottomrule
\end{tabular}
}
\caption{Comparison with interactive segmentation tools. Best and second-best results are shown in \textbf{bold} and \underline{underline}, respectively. $\times$ means that the model does not support text instruction following.}
\label{tab:toolperformance}
\end{table*}

\subsection{More Case Studies}
\label{sec:more_cases}

\begin{figure*}[t]
  \centering
  \includegraphics[width=1.0\linewidth]{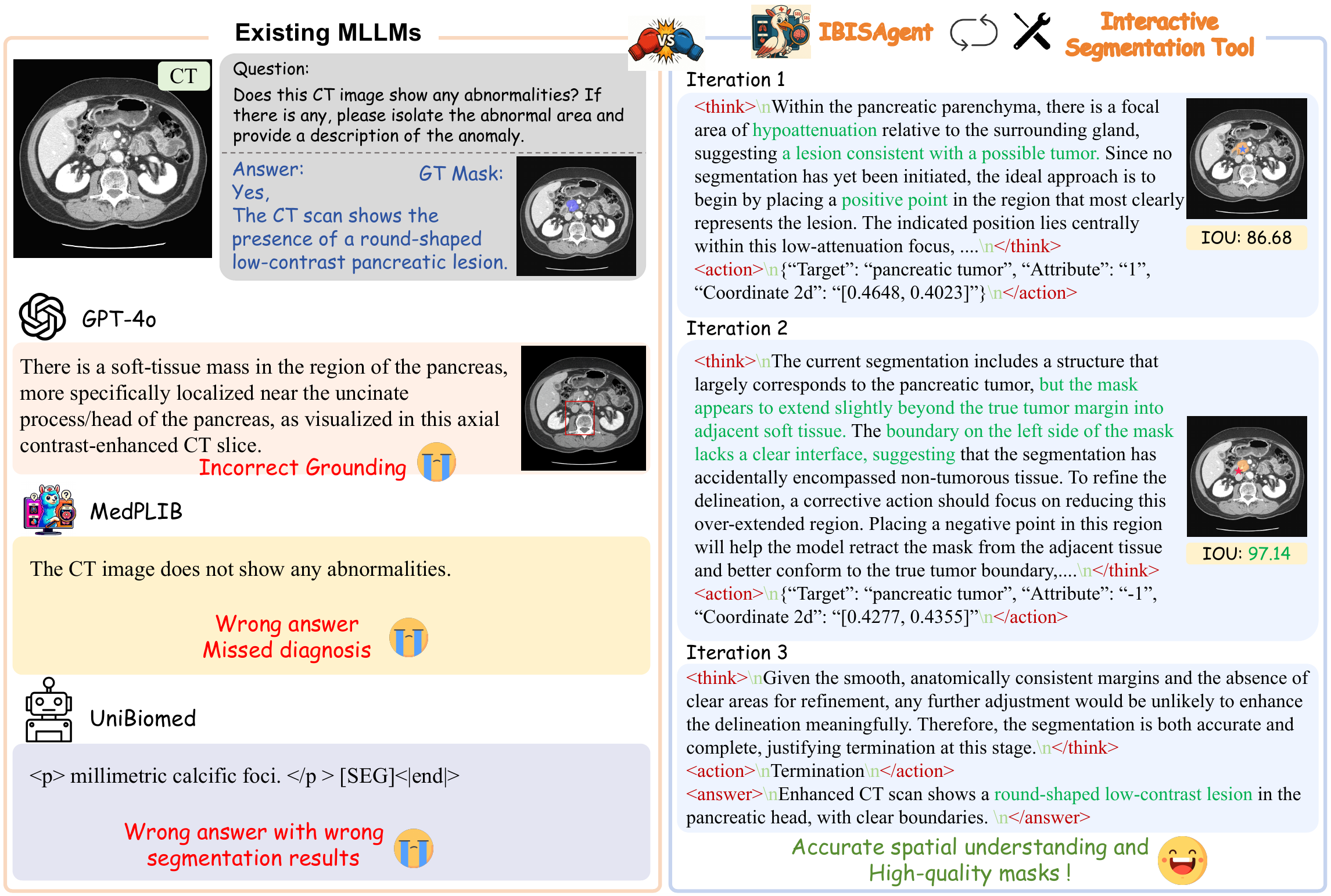} 
  \caption{
  \textbf{Qualitative comparison on the pancreatic tumor case.} 
  Existing MLLMs fail to provide reliable analysis: GPT-4o identifies the wrong location, MedPLIB misses the diagnosis entirely, and UniBiomed hallucinates unrelated calcific foci with an incorrect mask. 
  Conversely, IBISAgent accurately identifies the low-contrast lesion and performs multi-step refinement to distinguish the tumor from the surrounding pancreatic parenchyma, achieving an IoU of $97.14\%$.
  }
  \label{fig:pancreas_case}
\end{figure*}

\begin{figure*}[t]
  \centering
  \includegraphics[width=1.0\linewidth]{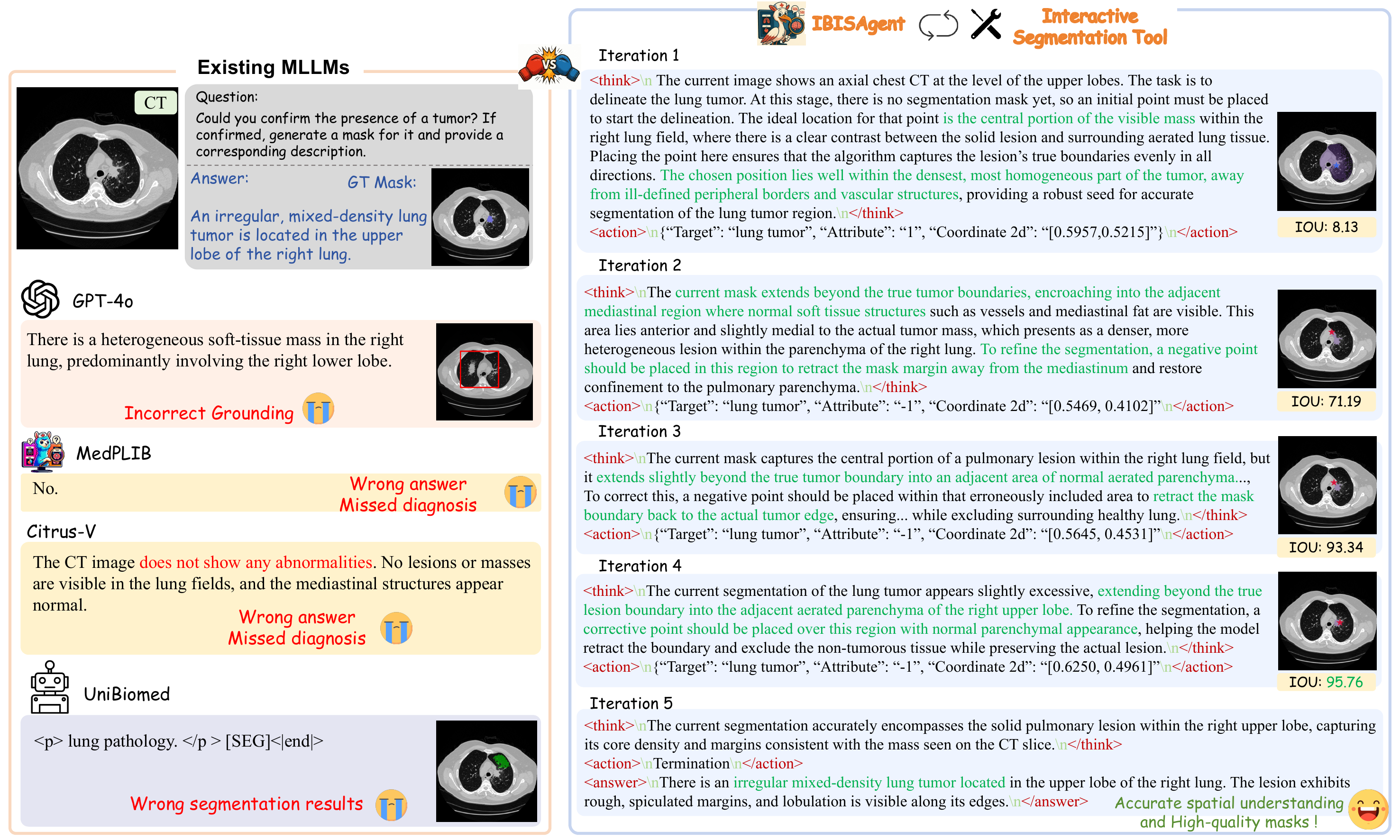} 
  \caption{
  \textbf{Qualitative comparison on the lung tumor case.} 
  While baseline models fail to detect the abnormality and GPT-4o mislocalizes the lesion, IBISAgent demonstrates superior pixel-level reasoning. 
  Through five iterations, the agent detects the irregular mass and progressively corrects over-segmentation errors in the mediastinal and parenchymal regions, resulting in a $95.76\%$ IoU.
  }
  \label{fig:lung_case}
\end{figure*}

We provide additional qualitative comparisons to further prove the robustness of IBISAgent in different anatomical regions. 
As illustrated in~\cref{fig:pancreas_case} and~\cref{fig:lung_case}, we present two challenging scenarios including a low-contrast pancreatic tumor and an irregular lung tumor, respectively. 
Consistent with our observations in~\cref{qualitative}, existing MLLMs frequently suffer from severe hallucinations, incorrect grounding, or missed diagnoses when discerning subtle pathological cues. 
In contrast, IBISAgent successfully initiates correct segmentation and employs its unique reasoning-driven refinement mechanism to correct errors—such as retracting masks from adjacent healthy tissues—ultimately achieving high-quality segmentation and accurate diagnostic descriptions.

\begin{figure*}[t]
  \centering
  \includegraphics[width=1.0\linewidth]{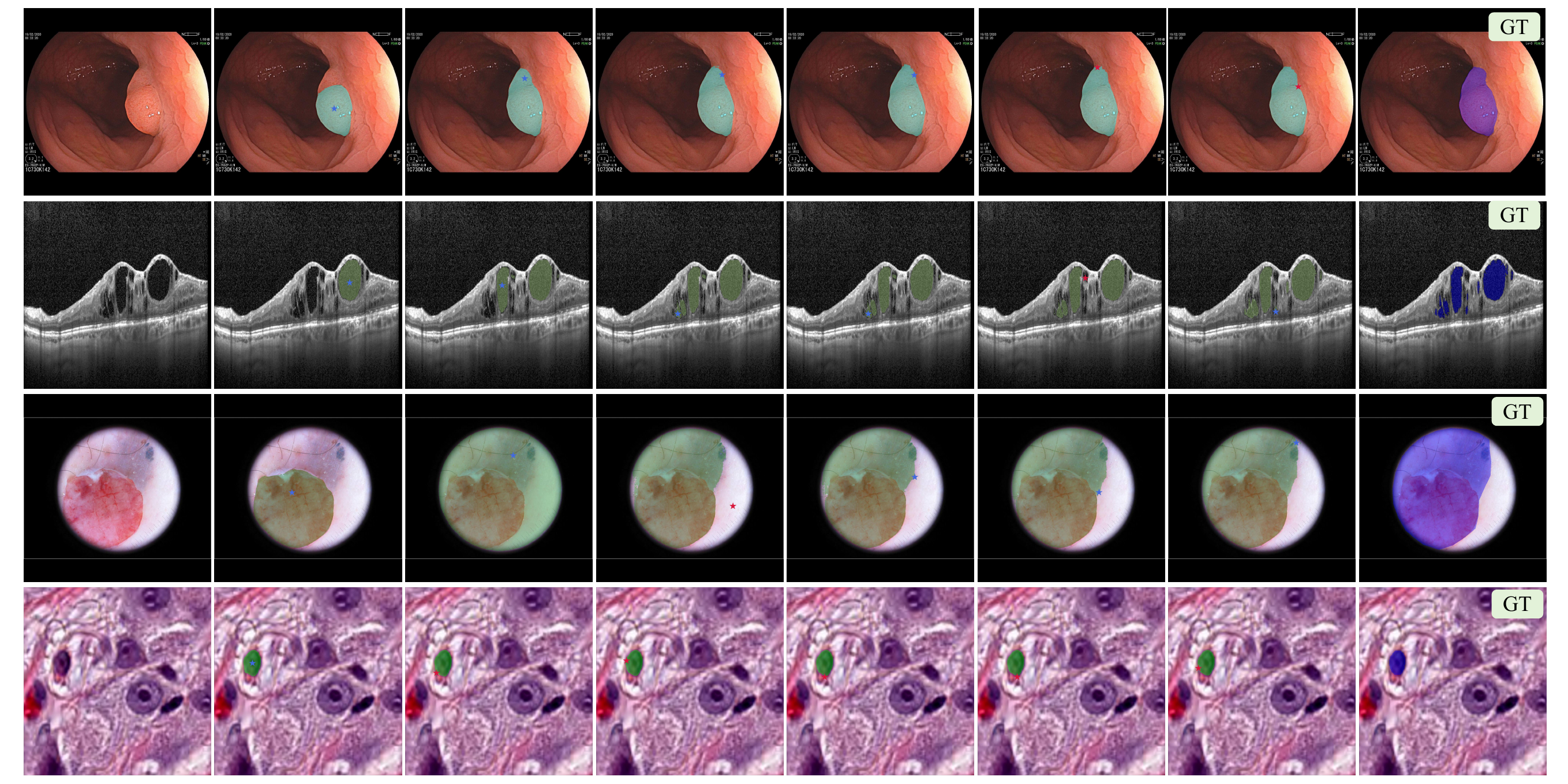} 
  \caption{Visualization of IBISAgent’s multi-round segmentation trajectories on various biomedical images, illustrating its iterative refinement process across different segmentation tasks.}
  \label{fig:trajectoryfigure}
\end{figure*}

We also present additional segmentation results in \cref{fig:trajectoryfigure}, showcasing IBISAgent’s multi-round segmentation trajectories on various biomedical images and illustrating its iterative refinement process across different segmentation tasks.

\subsection{How IBISAgent Corrects Errors}
\begin{figure*}[t]
  \centering
  \includegraphics[width=1.0\linewidth]{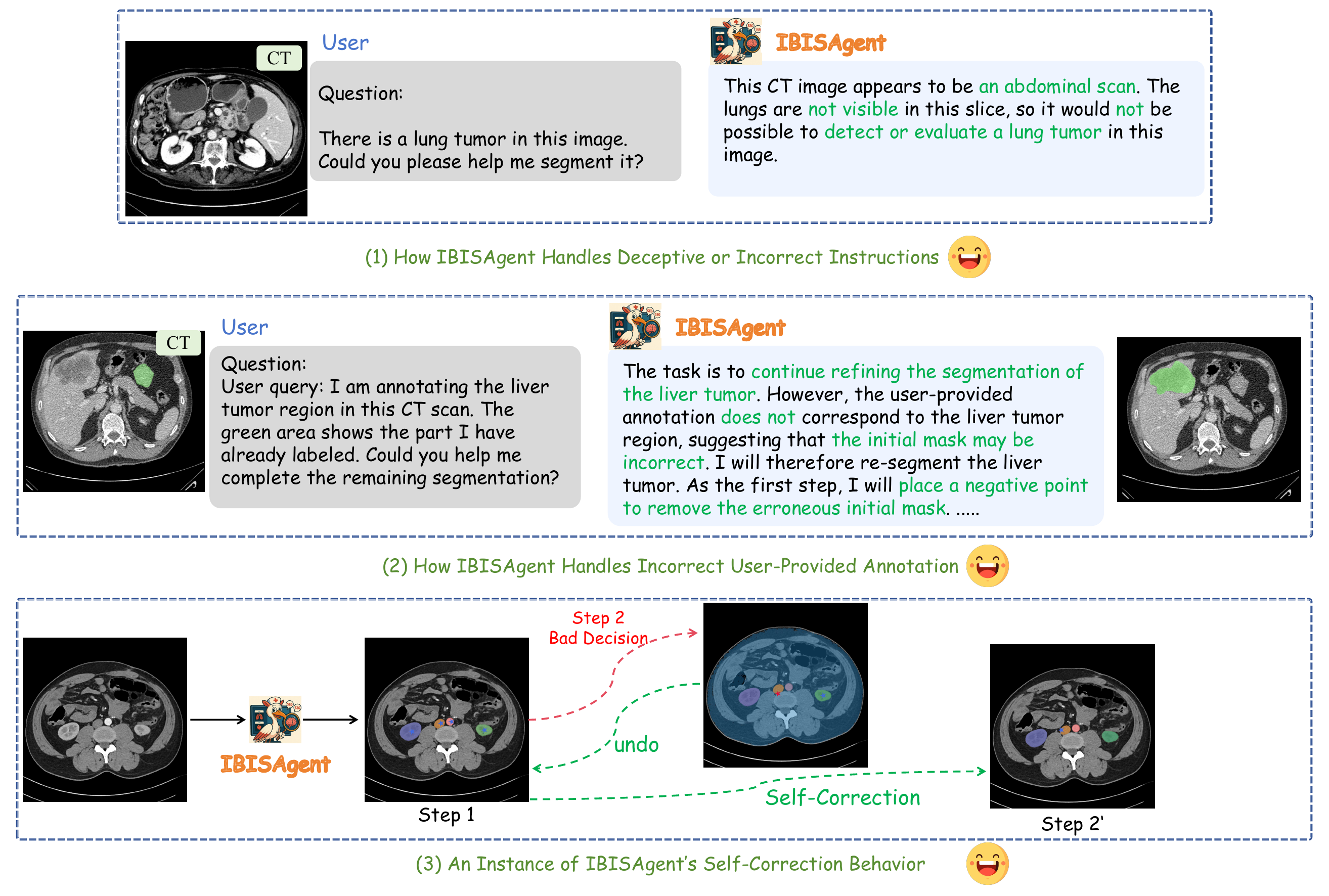} 
  \caption{\textbf{Illustrations of how IBISAgent corrects different types of errors.} (1) An example showing IBISAgent’s response when the user provides deceptive or incorrect instructions describing a nonexistent target. (2) A case where the initial mask provided by the user does not match the described segmentation target, and how IBISAgent reacts accordingly. (3) An example demonstrating IBISAgent’s ability to backtrack and undo an incorrect decision, followed by re-planning and selecting new click points.}
  \label{fig:SRB}
\end{figure*}

In~\cref{fig:SRB}, we present several representative examples that illustrate how IBISAgent corrects different types of errors, further demonstrating its robustness. We consider three typical scenarios.
(1) Deceptive or incorrect instructions: When the user provides misleading or erroneous instructions describing a nonexistent target, IBISAgent correctly recognizes that the specified object is not present in the image and refrains from producing an incorrect mask, highlighting its genuine understanding of fine-grained visual cues.
(2) Inconsistent initial masks: During mask refinement, if the user supplies an initial mask that does not match the described segmentation target, IBISAgent detects the inconsistency, corrects the erroneous mask, and generates the appropriate segmentation result.
(3) Backtracking to undo incorrect decisions: During multi-round mask refinement, IBISAgent has access to the full interaction history, enabling it to assess whether the current segmentation trajectory is reasonable. When an incorrect action leads to a suboptimal mask, the agent can automatically backtrack, undo the erroneous decision, and re-plan its click sequence. This ability largely stems from our use of Reflective Behavior Synthesis during SFT, which provides synthetic examples of such behavior and strengthens the agent’s robustness.

\subsection{RL Training Dynamics}

\begin{figure*}[t]
  \centering
  \includegraphics[width=1.0\linewidth]{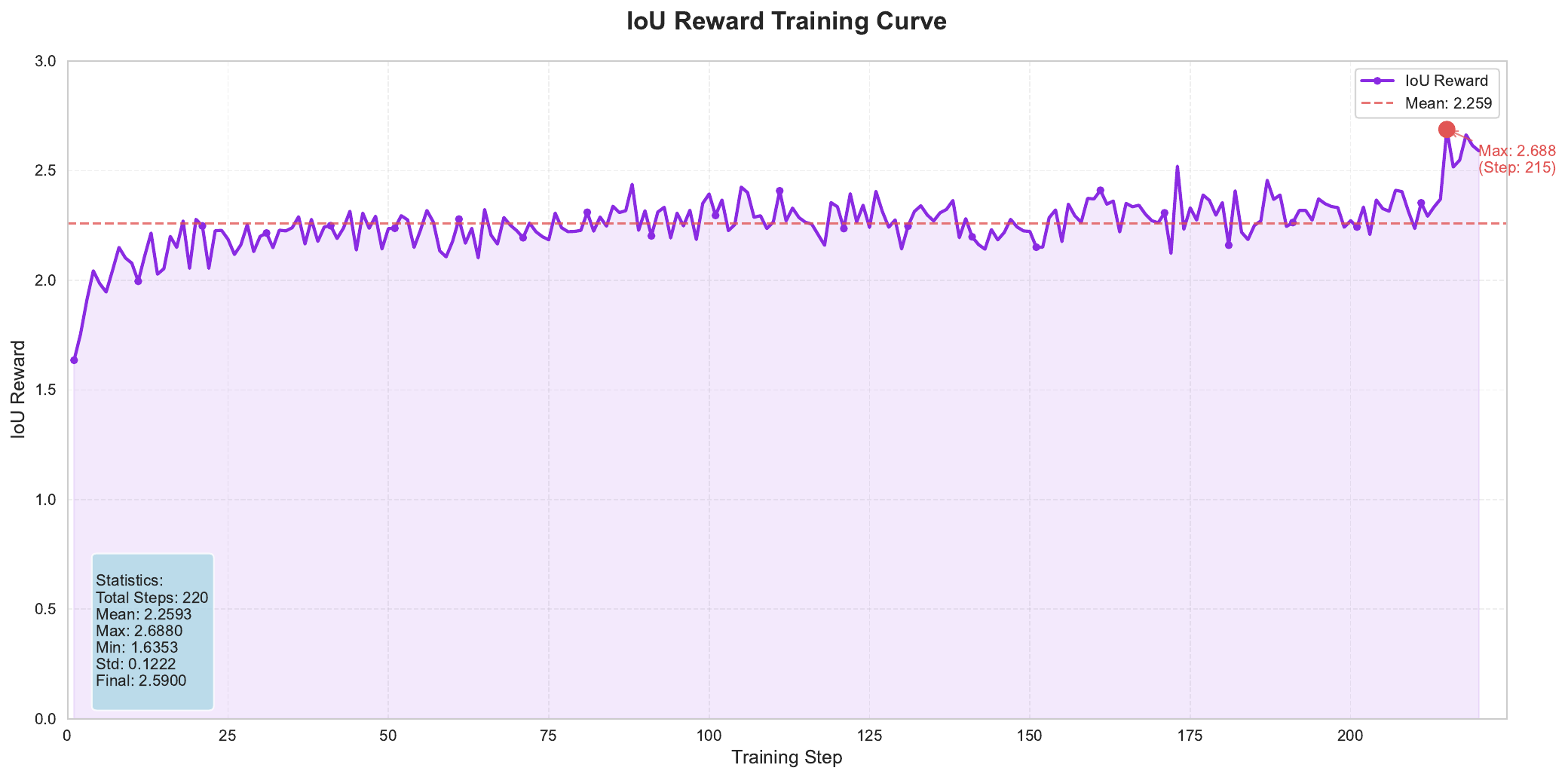}
  \caption{\textbf{The IoU reward curve.} We analyze the training dynamics to demonstrate that the RL training of IBISAgent show stable and consistent improvements.}
  \label{fig:IOUreward}
\end{figure*}

To analyze the training dynamics, we plot the IoU reward against training steps in~\cref{fig:IOUreward}. The curve illustrates the overall improvement in segmentation performance throughout the RL process. As observed, the IoU reward exhibits a steady increase, which demonstrates the stability of the training procedure. Under the guidance of our designed reward, the agent continuously explores the environment, thereby progressively acquiring enhanced planning and segmentation capabilities.

\section{System and User Prompts}
\label{sec:prompts}
\begin{figure*}[t]
    \centering
    \includegraphics[width=\linewidth]{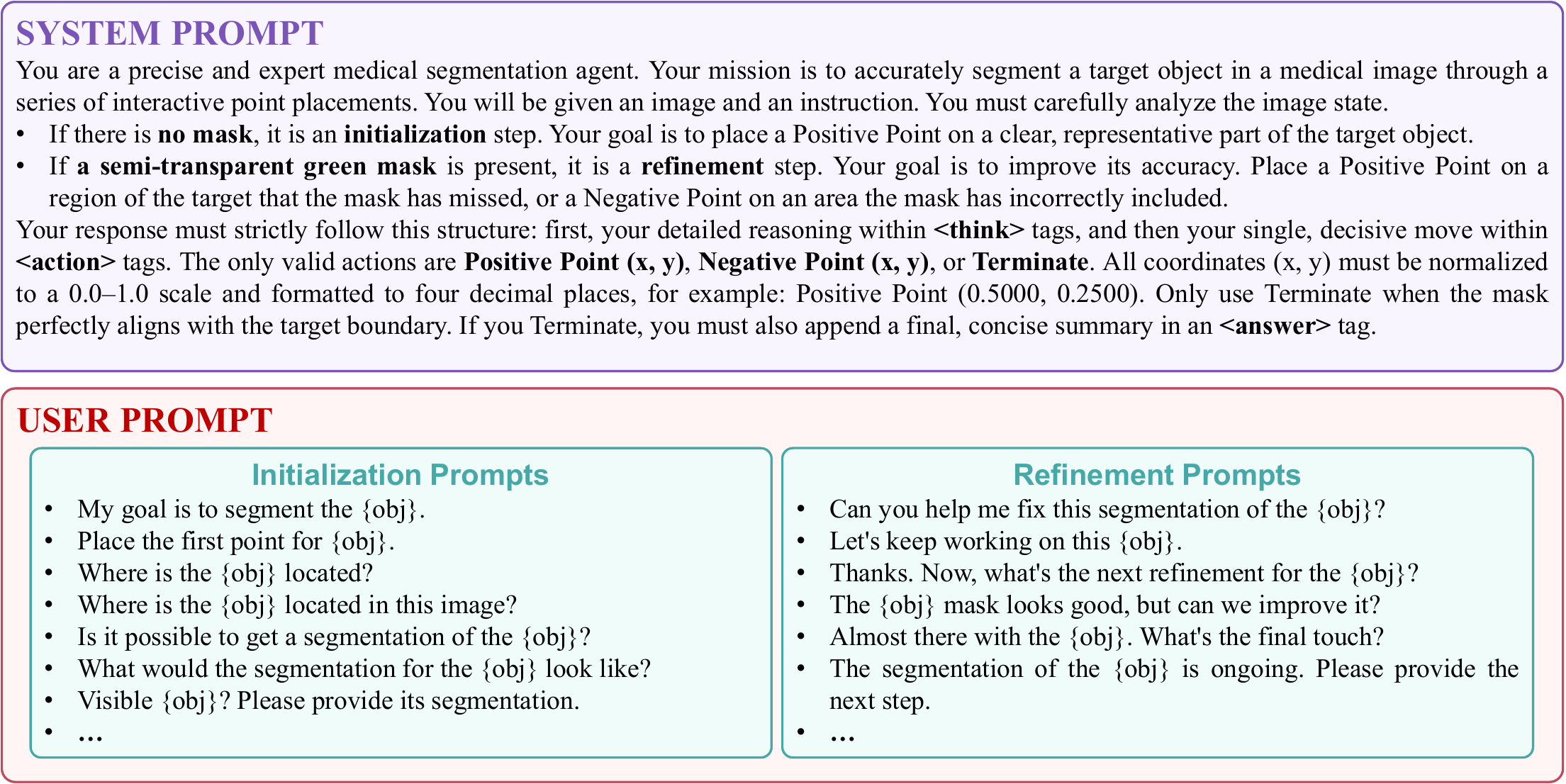}
    \caption{The system and user prompt used in IBISAgent. 
    }
    \label{fig:prompt}
\end{figure*}

The detailed system prompt and user prompt used by IBISAgent are shown as~\cref{fig:prompt}. 

\begin{table*}[p]  
  \centering
  \label{tab:sft_task_stats}
  \caption{Detailed statistics for the 38 task groups within our \textbf{SFT} dataset $\mathcal{D}_{cold}$.}
  \renewcommand{\arraystretch}{0.9}  
  \begin{tabular}{lccccccc}
    \toprule
    \textbf{Dataset (Group)} & \textbf{Samples} & \textbf{Avg. Length} & \textbf{Total QAs} & \textbf{Avg. IoU} & \textbf{Median IoU} & \textbf{Avg. DSC} & \textbf{Median DSC} \\
    \midrule
    ACDC                      &  1,746 &  8.60 &  16,769 & 0.9453 & 0.9459 & 0.9718 & 0.9722 \\
    BreastUS                  &    596 &  9.17 &   6,062 & 0.9351 & 0.9347 & 0.9664 & 0.9662 \\
    CAMUS                     &  1,996 &  9.66 &  21,278 & 0.9405 & 0.9388 & 0.9693 & 0.9684 \\
    COVID-19\_CT              &    224 &  9.08 &   2,258 & 0.8863 & 0.8826 & 0.9394 & 0.9376 \\
    COVID-QU\_Ex              &  1,854 &  8.82 &  18,207 & 0.9584 & 0.9586 & 0.9787 & 0.9789 \\
    CXR\_Masks\_and\_Labels     &  1,509 &  8.08 &  13,697 & 0.9470 & 0.9504 & 0.9727 & 0.9746 \\
    FH-PS-AOP                 &    878 &  9.63 &   9,336 & 0.9203 & 0.9180 & 0.9584 & 0.9572 \\
    G1020                     &    265 &  9.26 &   2,718 & 0.8834 & 0.8767 & 0.9377 & 0.9343 \\
    GlaS                      &    123 &  8.65 &   1,187 & 0.7368 & 0.7336 & 0.8422 & 0.8463 \\
    ISIC                      &  1,075 &  9.29 &  11,063 & 0.9331 & 0.9315 & 0.9653 & 0.9645 \\
    LGG                       &    910 &  9.05 &   9,148 & 0.9341 & 0.9345 & 0.9658 & 0.9662 \\
    LIDC-IDRI                 &    791 &  7.11 &   6,416 & 0.9254 & 0.9230 & 0.9612 & 0.9599 \\
    LiverUS                   &     29 &  9.21 &     296 & 0.8940 & 0.9188 & 0.9428 & 0.9577 \\
    MMs                       &  1,743 &  8.88 &  17,220 & 0.9484 & 0.9479 & 0.9735 & 0.9732 \\
    NeoPolyp                  &  1,473 &  5.96 &  10,252 & 0.9561 & 0.9611 & 0.9775 & 0.9802 \\
    OCT-CME                   &    205 &  8.02 &   1,850 & 0.9116 & 0.9074 & 0.9537 & 0.9515 \\
    PanNuke                   &  1,072 &  8.87 &  10,578 & 0.9278 & 0.9267 & 0.9625 & 0.9620 \\
    PolypGen                  &    958 &  6.50 &   7,182 & 0.9529 & 0.9603 & 0.9758 & 0.9797 \\
    QaTa-COV19                &    314 &  9.22 &   3,208 & 0.9049 & 0.8996 & 0.9500 & 0.9472 \\
    REFUGE                    &    207 &  9.30 &   2,132 & 0.8833 & 0.8776 & 0.9379 & 0.9348 \\
    Radiography               &  4,156 &  8.93 &  41,254 & 0.9480 & 0.9524 & 0.9732 & 0.9756 \\
    UWaterlooSkinCancer       &    227 &  7.91 &   2,023 & 0.9430 & 0.9455 & 0.9706 & 0.9720 \\
    amos22                    &  3,425 &  4.86 &  20,055 & 0.9640 & 0.9686 & 0.9816 & 0.9840 \\
    MSD brain tumor               &  1,925 &  5.94 &  13,366 & 0.9499 & 0.9490 & 0.9743 & 0.9738 \\
    MSD colon tumor               &    209 &  8.84 &   2,057 & 0.9138 & 0.9116 & 0.9549 & 0.9538 \\
    MSD heart                     &    812 & 15.59 &  13,473 & 0.9095 & 0.9144 & 0.9524 & 0.9553 \\
    MSD hepatic vessel            &    306 & 18.47 &   5,959 & 0.8453 & 0.8405 & 0.9158 & 0.9133 \\
    MSD hepatic vessel tumor      &     53 &  9.34 &     548 & 0.7469 & 0.7495 & 0.8543 & 0.8568 \\
    MSD hippocampus               &    319 & 19.91 &   6,669 & 0.9358 & 0.9231 & 0.9665 & 0.9600 \\
    kits23                    &  1,427 &  1.91 &   4,153 & 0.9712 & 0.9698 & 0.9854 & 0.9847 \\
    MSD liver                     & 10,034 &  6.95 &  79,816 & 0.9549 & 0.9532 & 0.9769 & 0.9760 \\
    MSD liver tumor               &    546 &  8.07 &   4,951 & 0.9313 & 0.9287 & 0.9643 & 0.9630 \\
    MSD lung tumor                &    299 &  6.90 &   2,361 & 0.9352 & 0.9347 & 0.9664 & 0.9663 \\
    MSD pancreas                  &  3,586 & 17.70 &  67,072 & 0.9240 & 0.9234 & 0.9604 & 0.9602 \\
    MSD pancreas tumor            &    474 &  7.50 &   4,030 & 0.9320 & 0.9314 & 0.9647 & 0.9645 \\
    MSD prostate                  &    204 & 19.16 &   4,113 & 0.8664 & 0.8601 & 0.9279 & 0.9248 \\
    siim-acr-pneumothorax     &    252 &  9.18 &   2,565 & 0.8658 & 0.8593 & 0.9278 & 0.9243 \\
    MSD spleen                    &    924 & 11.42 &  11,473 & 0.9458 & 0.9507 & 0.9721 & 0.9747 \\
    \midrule
    \textbf{Total}            & \textbf{47,146} & \textbf{8.69} & \textbf{456,795} & \textbf{0.9427} & \textbf{0.9507} & \textbf{0.9703} & \textbf{0.9747} \\
    \bottomrule
  \end{tabular}

\end{table*}

\section{Future Works}
IBISAgent endows MLLMs with powerful pixel-level visual reasoning capabilities. Together with our novel behavioral formulation and training framework, IBISAgent substantially pushes the boundary of MLLM-based biomedical image reasoning and segmentation. Nevertheless, several open challenges remain. First, the current agent operates primarily in 2D settings; extending IBISAgent to 3D scenarios—or even developing a unified 2D–3D MLLM agent—represents a highly promising research direction. Second, further improvements in the efficiency of multi-step agentic interaction will be essential to reduce computational overhead.

\end{document}